%% file: neurips_2026.tex
\documentclass{article}

\PassOptionsToPackage{numbers, compress}{natbib}
\usepackage[preprint]{neurips_2026}

\usepackage[utf8]{inputenc} 
\usepackage[T1]{fontenc}    
\usepackage{hyperref}       
\usepackage{url}            
\usepackage{booktabs}       
\usepackage{amsfonts}       
\usepackage{nicefrac}       
\usepackage{microtype}      
\usepackage{xcolor}         

\usepackage{algorithm}
\usepackage{algorithmic}
\usepackage{amsmath,amssymb}
\usepackage{amsthm}

\usepackage{booktabs}
\usepackage{multirow}
\usepackage{makecell}
\usepackage{colortbl}
\PassOptionsToPackage{table,xcdraw}{xcolor}

\usepackage{graphicx}
\usepackage{subfig}
\usepackage{wrapfig}
\usepackage{float}
\usepackage{pifont}
\usepackage[normalem]{ulem}
\useunder{\uline}{\ul}{}
\usepackage{mdframed}
\usepackage{xurl}

\definecolor{mycolor}{HTML}{F5F5F5}

\makeatletter
\newcommand{\equalcontrib}{\@fnsymbol{1}} 
\newcommand{\corresponding}{\@fnsymbol{2}} 
\makeatother

\title{ReSET: Accurate Latency-Critical NVFP4 Reasoning via Step-Aware Temperature Scaling}

\author{%
\textbf{Sihwa Lee}\textsuperscript{1\equalcontrib} \quad
\textbf{Janghwan Lee}\textsuperscript{1\equalcontrib} \quad
\textbf{Donghoon Yoo}\textsuperscript{2} \quad
\textbf{Jae Gon Kim}\textsuperscript{2} \\
\textbf{Hanyul Ryu}\textsuperscript{2} \quad
\textbf{Soojung Ryu}\textsuperscript{2} \quad
\textbf{Jungwook Choi}\textsuperscript{1\corresponding} \\
\textsuperscript{1}Hanyang University, Seoul, Republic of Korea \\
\textsuperscript{2}Xenoscube Korean Inc., Seoul, Republic of Korea \\
\texttt{\{macto94, hwanii0288\}@hanyang.ac.kr} \\
\texttt{\{donghoon.yoo, jay.kim, hanyul.ryu, sue.ryu\}@xenoscube.ai} \\
\texttt{choij@hanyang.ac.kr}
}

\begin{document}
\maketitle

\begingroup
\renewcommand\thefootnote{}
\footnotetext{\textsuperscript{\equalcontrib}Equal contribution.
\textsuperscript{\corresponding}Corresponding author.}
\endgroup

\vspace{-5mm}
\begin{abstract}
\vspace{-3mm}
Large reasoning models (LRMs) improve complex problem-solving by generating long intermediate reasoning traces, but this substantially increases inference costs. 
NVFP4 inference offers a promising approach to reduce both computational and memory costs through hardware-supported low-precision execution. 
However, directly applying NVFP4 to LRMs introduces two practical limitations: reasoning accuracy degrades under quantization, and existing NVFP4 kernels do not fully realize latency benefits in small-batch autoregressive decoding.
In this work, we analyze the effect of NVFP4 quantization on token-level uncertainty during reasoning. 
We show that quantization increases incorrect sampling at low-entropy symbolic tokens, while causing over-concentration on a small set of tokens in high-uncertainty reasoning steps.
Based on this observation, we propose \textbf{ReSET}, a reasoning-step entropy-based temperature-scaling method that estimates step-level uncertainty online and adapts the decoding temperature using both token-level and step-level entropy signals. 
To address the latency gap, we further design a CUDA-core small-$M$ NVFP4 kernel for latency-critical autoregressive decoding.
Across reasoning benchmarks and model scales, ReSET improves NVFP4 reasoning accuracy by up to $\sim\!$2 points over the NVFP4 baseline. Our CUDA-core small-$M$ kernel further improves latency-critical decoding, delivering up to $2.5\!\times$ kernel-level speedup over NVFP4 vLLM and approximately $2\!\times$ end-to-end decoding speedup over BF16. Code is available at \url{https://github.com/aiha-lab/ReSET}.
\end{abstract}

\input{Sections/introduction}
\input{Sections/background}

\input{Sections/observation}

\input{Sections/method}
\input{Sections/system}
\input{Sections/experiments}
\input{Sections/conclusion}

\bibliographystyle{unsrtnat}
\bibliography{refs}

\appendix
\input{Sections/appendix}

\newpage

\end{document}

%% file: Sections/introduction.tex
\section{Introduction}
\label{sec:introduction}

Large reasoning models (LRMs) have emerged as a powerful paradigm for solving complex multi-step mathematical and logical problems by generating extended chain-of-thought (CoT) traces~\cite{openai2024openaio1card,deepseek-r1,qwen3,openthoughts}.
Their accuracy gains rely on \emph{inference-time scaling} --- allocating additional compute during sequential decoding to handle harder instances --- which makes the dominant cost dimension \emph{per-token decode latency}, not aggregate throughput.
This cost grows along two compounding dimensions per generated token --- (i) memory bandwidth for repeated weight loading during autoregressive decoding~\cite{awq} and (ii) aggregate floating-point operations~\cite{atom} --- and is amplified by reasoning traces that routinely exceed 10K tokens under inference-time scaling~\cite{deepseek-r1-fp4}.

Of the levers available to compress these costs, low-precision execution directly targets these two axes at the hardware level: each low-precision element reduces both the bytes loaded per weight access and the per-element compute cost in proportion to the bit width.
NVFP4~\cite{b100}, the most recent low-precision format with native Blackwell Tensor-Core support, delivers $\sim$$4\times$ higher peak throughput than BF16 and a $\sim$$4\times$ smaller weight footprint, making it an attractive lever for low-latency LRM serving.

Realizing this benefit on LRMs, however, is non-trivial: two obstacles stand between NVFP4's headline numbers and a deployable system.

\textbf{Accuracy.} Reasoning involves intermediate symbolic tokens --- digits, operators, structural markers --- where small perturbations in token probabilities propagate through subsequent steps and break the final answer~\cite{wang2025beyond,lightman2024letsverify,liu2025quantizationhurts}.
Quantization-aware training mitigates this but is prohibitively expensive~\cite{lv2026makeslowbitquantizationawaretraining,nvidia-qad}; existing NVFP4 PTQ methods~\cite{duquantplus,fouroversix,blockrotation,mrgptq} target weight quantization error, which is already small at NVFP4's group size~\cite{lee-etal-2025-amxfp4,mrgptq} and does not correlate cleanly with reasoning accuracy.

\textbf{Latency.} The $4\times$ peak throughput above is delivered by Blackwell's \texttt{tcgen05.mma} instruction, whose tile is fixed at $M{=}128$ along the token dimension --- a structural requirement of Tensor-Core execution.
Production reasoning serving operates an order of magnitude below this: TPOT SLOs, KV-cache pressure, prefill--decode interference, and generation stalls~\cite{sarathi-serve,distserve,pagedattention} cap the SLO-feasible decode batch at $M\!\leq\!8$ on B200 (Sec.~\ref{sec:2.3}).
The result is below-$6.25\%$ tile occupancy and below-$1\%$ measured Tensor-Core utilization in the regime that governs production --- the headline $4\times$ advantage \emph{collapses} where it matters most.

We address both obstacles with a unified, deployment-oriented design that pairs a near-zero-cost decoding-time control policy with a kernel matched to the actual decode shape.
\textbf{ReSET} reads quantization-induced sampling errors at \emph{step-level} granularity --- the granularity at which reasoning-model uncertainty actually fluctuates --- and adapts decoding temperature accordingly, recovering up to $+2.6$ accuracy points on AIME-120 at $\sim$1.5\% per-decode-step overhead.
On the system side, we engineer a \textbf{CUDA-core NVFP4 GEMV kernel} for the small-$M$ decoding regime that Tensor-Core paths cannot serve, achieving $1.57$--$2.49\times$ projection latency reduction over vLLM-CUTLASS at $M{=}1$--$8$ on B200 and up to $1.97\times$ end-to-end latency reduction over BF16 on Qwen3-32B.
We are not aware of a prior public CUDA-core NVFP4 GEMV implementation; Code is available at \url{https://github.com/aiha-lab/ReSET}.

We make the following contributions:

\begin{itemize}
    \item We characterize NVFP4 reasoning failures at \emph{step-level} granularity (Sec.~\ref{sec:observations}), showing that token-level entropy --- the signal used by prior entropy-aware decoding --- is dominated by the uncertainty of the surrounding reasoning step and is therefore unreliable as a fixed-threshold control signal. We propose \textbf{ReSET} (Sec.~\ref{sec:method}), a step-aware temperature scaling policy whose per-token overhead is a constant-time scalar update.
    \item We design a CUDA-core small-$M$ NVFP4 GEMV kernel (Sec.~\ref{sec:implementation}) tailored to the latency-critical decoding regime where Tensor-Core paths are tile-bound, with no prior public implementation we are aware of.
    \item We demonstrate up to $+2.6$ accuracy points on AIME-120, $1.57$--$2.49\times$ kernel-level speedup over vLLM-CUTLASS at $M{=}1$--$8$, and up to $1.97\!\times$ E2E speedup over BF16 on B200 across five reasoning-capable open models (Sec.~\ref{sec:exp_settings}).
\end{itemize}

%% file: Sections/background.tex
\section{Background}

NVFP4 deployment for LRMs is bounded along two axes: reasoning accuracy under quantization (Sec.~\ref{sec:nvfp4_lrm}--\ref{sec:entropy_temp}) and decode-phase latency at small $M$ (Sec.~\ref{sec:2.3}). The first axis motivates the entropy-based decoding analysis of Sec.~\ref{sec:observations} and the policy of Sec.~\ref{sec:method}; the second axis motivates the kernel design of Sec.~\ref{sec:implementation}.

\subsection{NVFP4 Reasoning Models}
\label{sec:nvfp4_lrm}

NVFP4 is NVIDIA's microscaled FP4 format~\cite{rouhani2023microscalingdataformatsdeep,nvidia-nvfp4}, combining FP4 values with fine-grained shared scaling (group size 16, detailed configs are in Table~\ref{tab:mx_configs}). 
Recent hardware such as NVIDIA B200~\cite{b100} natively supports NVFP4 Tensor Cores, achieving up to 9 PFLOPs and $\sim$4$\times$ higher throughput than BF16.
NVIDIA’s ModelOPT~\cite{modelopt} framework further enables practical deployment of NVFP4 reasoning models~\cite{openai2025gptoss120bgptoss20bmodel,nemotron-nano-nvfp4}.
However, applying FP4 quantization to LRMs remains challenging because quantization errors can accumulate over long reasoning trajectories~\cite{liu2025quantizationhurts}. 
Prior work addresses this through either training-based approaches, such as quantization-aware training and distillation~\cite{lv2026makeslowbitquantizationawaretraining,nvidia-qad}, or post-training quantization (PTQ) methods that reduce quantization error through techniques such as rotation, channel-wise scaling, and dynamic block-scale selection~\cite{duquantplus,blockrotation,mrgptq,fouroversix}.
However, PTQ methods can introduce additional runtime overhead due to online rotation for activations, and their gains often diminish for fine-grained formats such as NVFP4 where quantization error is already relatively small~\cite{mrgptq,nvidia-nvfp4}. 
This suggests that improving NVFP4 reasoning quality may require going beyond quantization-error minimization and considering the decoding process itself.

\subsection{Token Entropy and Temperature under Quantized Reasoning Decoding}
\label{sec:entropy_temp}

\textbf{Token entropy in reasoning.}
Token-level uncertainty in LRMs is heterogeneous and is quantified by token entropy: $H_t = -\sum_{v \in \mathcal{V}} p_t(v)\log p_t(v),$
where $p_t(v)$ denotes the probability of token $v$ at step $t$~\cite{beyond}.
Quantization perturbs this distribution~\cite{qdpo,qerl}: the resulting entropy shifts can encourage exploration during RL~\cite{qerl} but distort sampling at inference~\cite{qdpo}. \emph{Where} these shifts land --- and how they should be controlled --- is the question Sec.~\ref{sec:observations} takes up.

\textbf{Decoding temperature in LRMs.}
Decoding temperature controls the sharpness of the token distribution through $\mathrm{softmax}(z_t/T)$.
LRMs typically use non-zero temperatures (e.g., 0.6) because greedy decoding degrades reasoning accuracy~\cite{qwen3,deepseek-r1}.
While temperature scaling has been studied for test-time reasoning diversity~\cite{ontheroleoftemperature}, its interaction with quantized inference remains underexplored.

\begin{wrapfigure}{r}{0.42\textwidth}
    \centering
    \includegraphics[width=0.43\textwidth]{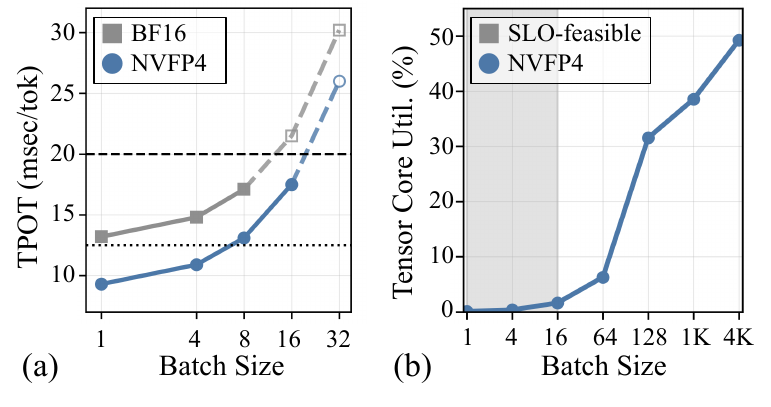}
    \vspace{-2mm}
    \caption{Limited batch-size scaling in Qwen3-32B on a single B200.
    (a) TPOT at 32K context; (b) Tensor Core utilization of NVFP4 decode GEMMs.}
    \label{fig:motivation-small-batch}
    \vspace{-4mm}
\end{wrapfigure}
\subsection{Latency-Critical Large Reasoning Model Serving}
\label{sec:2.3}
\textbf{The throughput--latency gap.}
NVFP4's headline 4$\times$ peak throughput over BF16~\cite{b100} is realized at compute-bound batch sizes, but production LRM serving operates an order of magnitude below that point.
LRM workloads produce long outputs --- R1~\cite{deepseek-r1} generates on average $\sim$12K tokens per AIME~\cite{aime} problem, up to 64K~\cite{deepseek-r1,openai2024openaio1card} --- so per-token decode latency, not aggregate throughput, sets the SLO~\cite{sarathi-serve,distserve}.

\textbf{The small-$M$ utilization collapse.}
Fig.~\ref{fig:motivation-small-batch}(a) reports TPOT as a function of decode batch size $M$ for Qwen3-32B at 32K context on a single B200; representative SLOs~\cite{adaserve} are crossed at moderate batch sizes, confining feasible operation to $M\!\leq\!8$.
\emph{In this region, NVFP4 Tensor-Core utilization stays below 1\%} (Fig.~\ref{fig:motivation-small-batch}(b)) --- a near-total collapse of the peak-throughput advantage NVFP4 was deployed for.
Prior FP4 quantization studies operate at the throughput-oriented batch sizes where this collapse is hidden; the latency-critical SLO-feasible regime, the focus of this paper, sits where NVFP4's headline benefit goes missing.
Sec.~\ref{sec:implementation} explains the mechanism --- fixed Tensor-Core tile granularity --- and engineers the matching kernel.

%% file: Sections/observation.tex
\vspace{-2mm}
\section{Observations}
\label{sec:observations}
\vspace{-2mm}
We probe NVFP4 reasoning failures along the granularity at which existing decoding methods reason about uncertainty: \emph{token-level entropy}~\cite{wang2025beyond}. Under W4A4, mis-sampling concentrates at low token-level-entropy symbolic tokens (Sec.~\ref{sec:3.1}) and a token-level entropy threshold partially recovers accuracy (Sec.~\ref{sec:3.2}), but whether a token \emph{appears} low- or high-entropy is dominated by the uncertainty of the surrounding reasoning step (Sec.~\ref{sec:3.3}). We therefore develop a \emph{step-level entropy} analysis and use it as the primary control signal for quantized reasoning.

\subsection{Low-Entropy Tokens as Quantization-Sensitive Decisions}
\label{sec:3.1}

\begin{figure}[t]
\centering
\centerline{\includegraphics[width=\columnwidth]{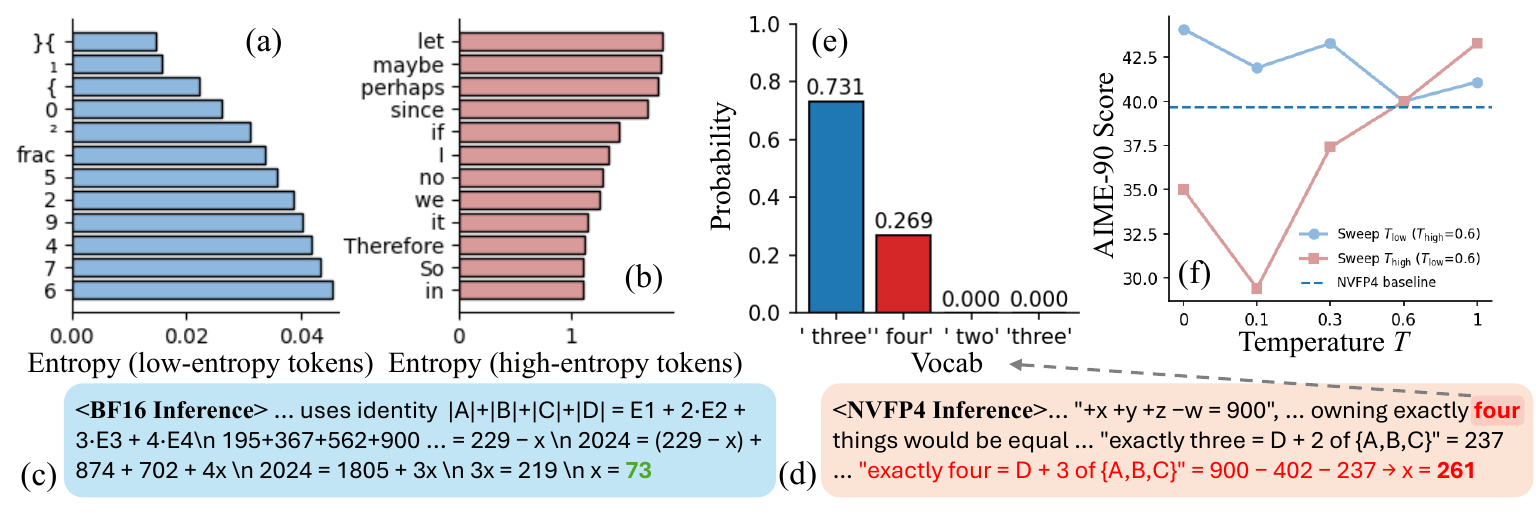}}
\vspace{-2mm}
\caption{
Example of (a) low- and (b) high-entropy tokens.
Example reasoning under (c) BF16 and (d) NVFP4 on R1-Qwen-7B.
(e) Next-token probability distribution at the symbolic position in (d).
(f) Effect of entropy-conditioned temperature control accuracy (R1-Qwen-7B, AIME 2022-2024).
}
\vspace{-3mm}
\label{fig:entropy_overview}
\end{figure}

We categorize 1.5M tokens generated from the R1-Qwen-14B~\cite{deepseek-r1} model by their entropy, following~\cite{wang2025beyond} (detailed analysis setting is in Appendix~\ref{appendix:token_analysis_settings}).
As shown in Fig.~\ref{fig:entropy_overview}(a-b), low-entropy tokens correspond to locally constrained \textit{symbolic decisions} such as digits and operators, while high-entropy tokens appear at semantically flexible \textit{branching points}. As described in Fig.~\ref{fig:entropy_overview}(c-d), under NVFP4 quantization, errors at symbolic tokens (e.g., ``four'') can propagate through subsequent reasoning and lead to incorrect final outcomes.
Notably, these errors are not driven by top-1 flips. As illustrated in Fig.~\ref{fig:entropy_overview}(e), the correct token (``three'') remains the most likely, but a non-top-1 alternative receives enough probability mass to be occasionally sampled, breaking the symbolic continuation.

\subsection{A Naive Fix: Entropy-Thresholded Temperature Sharpening}
\label{sec:3.2}

This suggests a direct intervention: sharpening the distribution at low-entropy positions to suppress mis-sampling. We test it with a fixed-threshold rule that, at each decoding position, computes the token entropy $H_t$, partitions tokens by an entropy threshold $\tau_0$, and applies a different temperature to each group:
\begin{equation}
\label{eq:fixed_threshold_temperature}
    T_t =
    \begin{cases}
    T_{\mathrm{low}}, & H_t < \tau_0, \\
    T_{\mathrm{high}}, & H_t \ge \tau_0.
    \end{cases}
\end{equation}
Here, $\tau_0$ is set to approximately $0.6$, corresponding to the top 20\% of token entropies observed in model responses.
As shown in Fig.~\ref{fig:entropy_overview}(f), this asymmetric sharpening improves NVFP4 reasoning accuracy: lowering the temperature $T_{\mathrm{low}}$ for low-entropy tokens yields consistent gains, while changes to the high-entropy temperature provide little benefit beyond moderate increases in $T_{\mathrm{high}}$.

\subsection{Token Entropy is Dominated by Step-Level Uncertainty}
\label{sec:3.3}

The result in Sec.~\ref{sec:3.1} is positive but not yet operational: it shows that \emph{some} notion of ``low-entropy token'' is the right intervention target, but the fixed-threshold rule's incomplete recovery indicates that token-level entropy is not the right signal.
This limitation stems from a granularity mismatch: prior entropy-aware decoding methods reason about uncertainty at the \emph{token} level, but the uncertainty that actually fluctuates during reasoning lives at the \emph{step} level.
We find that token-level entropy is not determined solely by the local token identity but is strongly influenced by the uncertainty of the surrounding reasoning step.
Even typically deterministic tokens (e.g., digits) can exhibit elevated entropy when they appear in uncertain reasoning steps, which can in turn increase the likelihood of incorrect symbolic decisions.
To analyze this effect, we group reasoning traces into step-level segments and examine how low- and high-token-level-entropy groups (Fig.~\ref{fig:entropy_overview}(a-b)) behave within each step.

\begin{figure}[t]
\centering
\centerline{\includegraphics[width=\columnwidth]{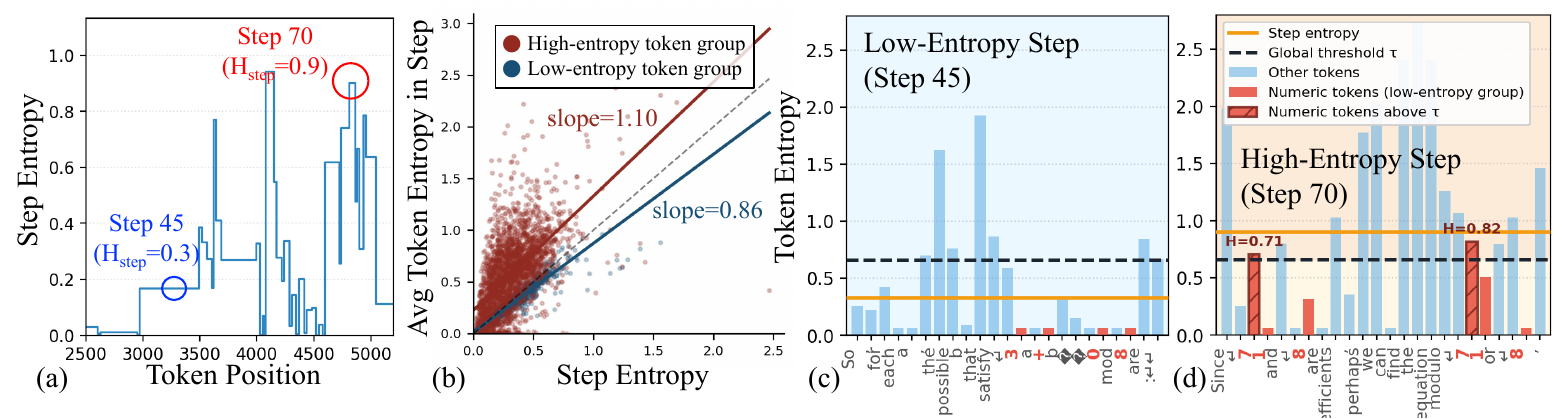}}
\vspace{-2mm}
\caption{
Step-level entropy dynamics.
(a) Step-wise entropy trajectory. 
Relationship between step entropy and token entropy for (b) low- and high-entropy groups. 
Token entropy distributions in representative (c) low- and (d) high-entropy steps.
}
\vspace{-3mm}
\label{fig:entropy_step}
\end{figure}

\textbf{Definition of a reasoning step.}
We define a \emph{reasoning step} as a contiguous segment of tokens corresponding to a coherent intermediate reasoning unit. 
Let $\{1,\dots,T\}$ denote the token positions in a sequence, partitioned into segments $\{\mathcal{S}_k\}_{k=1}^{K}$. 
For a token position $t \in \mathcal{S}_k$, we define the \emph{step entropy} as the average token entropy within the current reasoning step:
$H_{\mathrm{step}}(t)=\frac{1}{|\mathcal{S}_k|}\sum_{i\in\mathcal{S}_k} H_i$.

\textbf{Observations.}
Fig.~\ref{fig:entropy_step}(a) shows that step entropy varies substantially over the course of a single reasoning trace.
For example, Step 45 has low step entropy ($H_{\mathrm{step}}\!=\!0.3$), while Step 70 has high step entropy ($H_{\mathrm{step}}\!=\!0.9$), indicating that the model alternates between confident and uncertain reasoning states during generation --- variation that token-level entropy alone does not capture.
Fig.~\ref{fig:entropy_step}(b) makes the token--step coupling concrete: the mean token-level entropy of both the low- and high-entropy token groups closely tracks the step entropy of the segment they fall in. In other words, the same token category occupies different absolute positions on the token-level entropy axis depending on the surrounding step's uncertainty, and even deterministic symbolic tokens can exhibit elevated token-level entropy inside an uncertain reasoning step.

The practical consequence is illustrated in Fig.~\ref{fig:entropy_step}(c-d).
In low-entropy steps (Fig.~\ref{fig:entropy_step}(c)), symbolic tokens fall below the global threshold $\tau_0$, and a fixed-threshold rule correctly identifies them as positions to sharpen.
However, in high-entropy steps (Fig.~\ref{fig:entropy_step}(d)), symbolic tokens can exceed the same threshold due to elevated step-level uncertainty.
These tokens remain locally deterministic, but a token-level rule misclassifies them as high-entropy positions, causing the decoding rule to skip necessary sharpening.
The diagnosis is therefore not that the symbolic-token signal is wrong, but that it must be read \emph{relative to the current step} --- motivating a shift from token-level to step-level entropy analysis. We operationalize this step-relative formulation, which combines step-level uncertainty with token-level deviation within a step, in Sec.~\ref{sec:method}.

%% file: Sections/method.tex
\vspace{-1mm}
\section{ReSET: Reasoning Step Entropy-based Temperature Scaling}
\label{sec:method}
\vspace{-1mm}
We realize the token-to-step shift of Sec.~\ref{sec:observations} as \textbf{ReSET}, which replaces the fixed token-level threshold of Eq.~\ref{eq:fixed_threshold_temperature} with one that adapts to step-level uncertainty. ReSET has two components: a \emph{step-aware threshold} that is global in confident steps and step-relative in uncertain ones, and an \emph{online step-entropy estimator} that supplies the per-token statistic this threshold needs under the causality of autoregressive decoding. The latency-side realization is deferred to Sec.~\ref{sec:implementation}.

\textbf{Step-aware threshold (SAT).}
The fixed-threshold rule of Sec.~\ref{sec:3.2} fails in exactly one regime: high-uncertainty steps, where step-level entropy lifts symbolic tokens above the global cutoff $\tau_0$ (Fig.~\ref{fig:entropy_step}(d)). We target this regime by comparing the online step-entropy estimate $\hat{H}_{\mathrm{step}}(t)$ (defined below) against the running global mean $\bar{H} \!=\! \frac{1}{t+1}\sum_{i=0}^{t} H_i$:
\begin{equation}
\label{eq:sat}
\tau_t =
\begin{cases}
\tau_0, & \hat{H}_{\mathrm{step}}(t) \le \bar{H} \quad\text{(confident step)}, \\
\hat{H}_{\mathrm{step}}(t), & \hat{H}_{\mathrm{step}}(t) > \bar{H} \quad\text{(uncertain step)},
\end{cases}
\end{equation}
where $\tau_0$ is the global threshold of Sec.~\ref{sec:3.2}; substituting Eq.~\ref{eq:sat} into Eq.~\ref{eq:fixed_threshold_temperature} gives the full ReSET policy.

\begin{figure}[t]
\centering
\centerline{\includegraphics[width=\columnwidth]{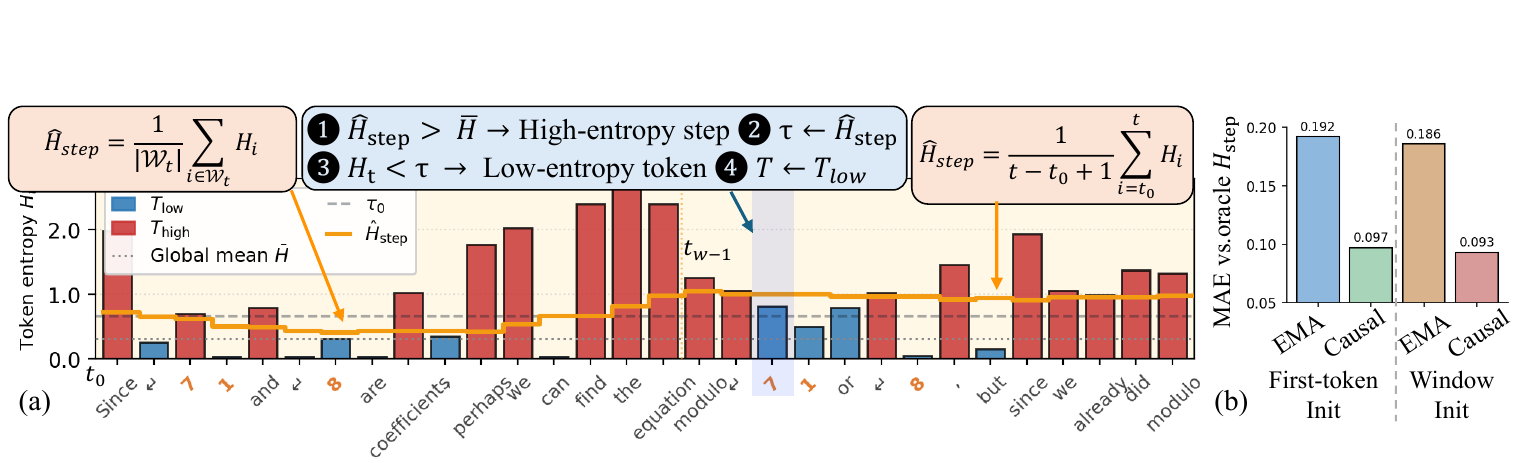}}
\vspace{-2mm}
\caption{
(a) ReSET temperature assignment on an R1-Qwen-14B AIME-120 trace.
(b) Mean absolute error of online step-entropy estimators against the offline oracle of Sec.~\ref{sec:3.3}.
}
\vspace{-4mm}
\label{fig:reset}
\end{figure}

The novelty lies in the uncertain branch: setting $\tau_t \!=\! \hat{H}_{\mathrm{step}}(t)$ makes the rule \emph{step-relative} --- a token is sharpened iff its token-level entropy is below the typical token of its own step, regardless of where the global cutoff sits. This recovers the symbolic-vs-branching distinction inside a step that Sec.~\ref{sec:3.3} showed token-level rules collapse: in Fig.~\ref{fig:reset}(a), the symbol ``7'' crosses $\tau_0$ but stays below $\hat{H}_{\mathrm{step}}(t)$, and SAT correctly sharpens it. The choice introduces no additional sweepable hyperparameter; $\tau_t$ is fully determined by quantities already tracked. In confident steps $\hat{H}_{\mathrm{step}}(t)$ becomes degenerately small, and a step-relative cutoff would misfire on symbolic tokens (Fig.~\ref{fig:fig_hres_failure}); we therefore fall back to $\tau_0$, which Sec.~\ref{sec:3.2} validated for this regime.

\textbf{Online step-entropy estimation.}
SAT needs $\hat{H}_{\mathrm{step}}(t)$ at every decoding position, but the offline definition of Sec.~\ref{sec:3.3} averages over not-yet-generated tokens. A causal estimator must therefore handle two within-step regimes with opposite bias-variance trade-offs.

\emph{Late regime: within-step aggregation.}
Once enough samples within the step are available, the lowest-bias estimate is the within-step running average $\hat{H}_{\mathrm{causal}}(t)\!=\!\tfrac{1}{t - t_0 + 1}\sum_{i=t_0}^{t} H_i$, where $t_0$ is the step's start. As a baseline, we use an exponential moving average (EMA; Appendix Eq.~\ref{eq:step_entropy_estimators}), which weights recent tokens but ignores the step boundary. Fig.~\ref{fig:reset}(b) shows the running average wins consistently --- the step boundary, not recency, is the dominant signal.

\emph{Early regime: step-transition initialization.}
In the first $w$ positions of a step, the within-step average is high-variance: few samples, and early samples are often atypical (step-opening discourse tokens). A sliding-window initialization over the $w$ most recent tokens (spanning the boundary) trades a small step-boundary bias for substantially lower variance until the step is self-supporting (Fig.~\ref{fig:reset}(b)).

The hybrid estimator HSE switches between the two regimes:
\begin{equation}
\label{eq:hstep_estimate}
\hat{H}_{\mathrm{step}}(t) =
\begin{cases}
\frac{1}{|\mathcal{W}_t|} \sum_{i \in \mathcal{W}_t} H_i, & t - t_0 < w \quad\text{(initialization)}, \\
\frac{1}{t - t_0 + 1} \sum_{i=t_0}^{t} H_i, & t - t_0 \ge w \quad\text{(within-step)},
\end{cases}
\end{equation}
with $\mathcal{W}_t$ the size-$w$ window ending at $t$.
HSE adds only lightweight per-token bookkeeping---one entropy reduction, two scalar updates ($\hat{H}_{\mathrm{step}}$, $\bar{H}$), and one branch---which is negligible compared with a full decode step and is included in the end-to-end measurements in Sec.~\ref{exp:e2e}.

\textbf{Decoding temperatures and calibration.}
We use $T_{\mathrm{low}} \!<\! T_{\mathrm{base}} \!<\! T_{\mathrm{high}}$ around the model's recommended base ($T_{\mathrm{base}}{=}0.6$). $T_{\mathrm{high}}$ is method-relevant: NVFP4 \emph{compresses} the entropy of high-entropy tokens in uncertain steps (e.g., ``Alternatively''; Fig.~\ref{fig:threshold_comparison}), so a temperature above the default is needed to restore the diversity quantization removes; empirically, $T_{\mathrm{high}}{=}1.0$ consistently improves accuracy (Table~\ref{tab:thigh_sweep_aime120}). $T_{\mathrm{low}}$ is selected per (model, task) on a held-out calibration split disjoint from evaluation (Table~\ref{tab:tlow_sweep}); $\tau_0$ is the 80th-percentile token entropy on the same split. We use $w{=}32$ (Table~\ref{tab:window_size}).

%% file: Sections/system.tex
\section{NVFP4 Kernel Design for Small-$M$ Decoding}
\label{sec:implementation}
\vspace{-2mm}
ReSET recovers reasoning quality, but realizing NVFP4's latency benefit hinges on a decode-phase kernel matched to the small-$M$ regime motivated in Sec.~\ref{sec:2.3}.
We therefore propose a CUDA-core NVFP4 kernel for the linear projection $Y{=}XW$ ($X\!\in\!\mathbb{R}^{M\!\times\! K}$, $W\!\in\!\mathbb{R}^{K\!\times\! N}$) at small $M$, and dispatch to a Tensor-Core path for prefill and large-$M$ decode (Appendix~\ref{sec:appendix-gemm}).
\vspace{-1mm}
\subsection{Tensor-Core Inefficiency at Small $M$}

\begin{figure}[t]
\centering
\centerline{\includegraphics[width=\columnwidth]{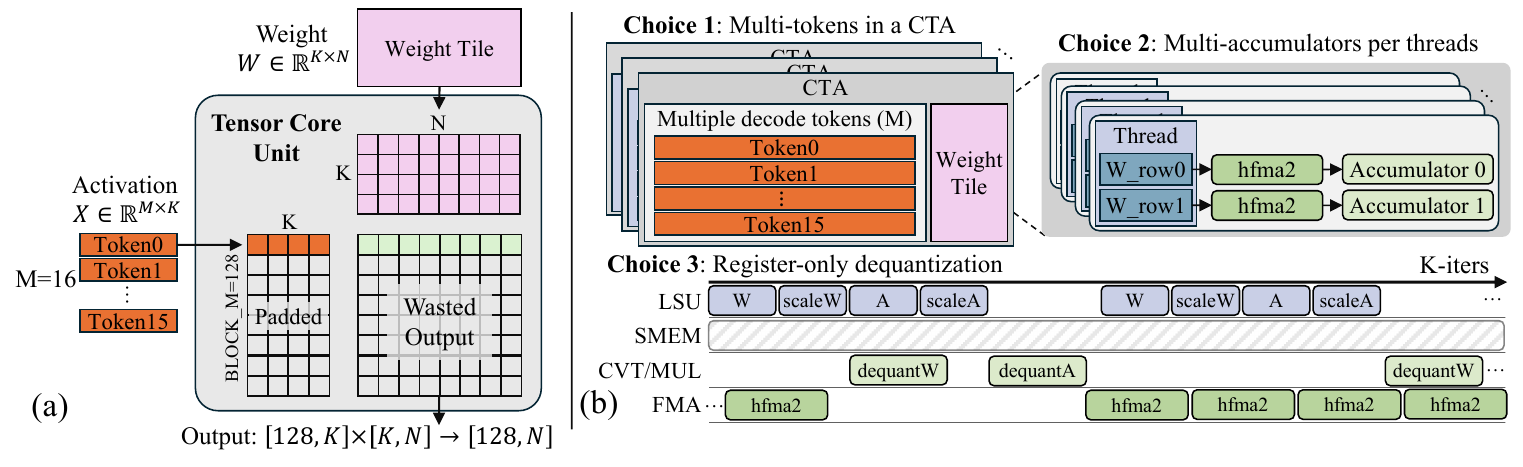}}
\vspace{-2mm}
\caption{
CUDA-core NVFP4 kernel design for small-$M$ decoding.
(a) Tensor-Core GEMM underutilization from the required 128-row $M$ tile.
(b) CUDA-core design choices.
}
\label{fig:cuda-core-kernel}
\vspace{-5mm}
\end{figure}

\textbf{Tile under-occupancy explains the Sec.~\ref{sec:2.3} collapse.}
The sub-$1\%$ utilization of Sec.~\ref{sec:2.3} traces to a fixed-tile constraint of Blackwell's NVFP4 GEMM path. Existing stacks (vLLM, CUTLASS, MR-GPTQ~\cite{mrgptq}) dispatch linear layers to utilize \texttt{tcgen05.mma}, whose tile is fixed at $M{=}128$ along the token dimension (Fig.~\ref{fig:cuda-core-kernel}(a)), so the logical $M\!\times\! K$ activation is padded into $128\!\times\! K$ and only $M$ of $128$ output rows are useful: $6.25\%$ tile occupancy at $M{=}8$, $3.13\%$ at $M{=}4$.
Throughput-oriented stacks hide this with batch sizes in the thousands; in the SLO-feasible region, the advantage is lost.

\textbf{Missing NVFP4 CUDA-core kernel.}
A CUDA-core path is well matched to small-$M$ decoding because it exposes $M$ flexibly at the thread level rather than through a fixed tile.
However, current NVFP4 inference frameworks continue to rely on Tensor-Core GEMM paths for decode projections, leaving this CUDA-core small-$M$ regime unsupported.
We therefore design, to our knowledge, the first CUDA-core NVFP4 GEMV kernel tailored to latency-critical small-$M$ decoding, with support for NVFP4's E4M3 block scales and FP4 unpacking pipeline.

\textbf{Challenges.}
A CUDA-core NVFP4 path must replace work that Tensor-Core MMA hides implicitly: (C1) reusing streamed weight tiles across active decode tokens, since every token shares $W$;
(C2) launching enough parallel threads with independent accumulator chains, since a naive mapping serializes \texttt{half2} FMAs along $K$ and under-occupies CUDA cores at small $M$; and
(C3) performing FP4 unpacking and shared-scale dequantization without materializing intermediate buffers or inserting synchronization in the inner $K$ loop.

\subsection{CUDA-Core Decode Kernel}
\label{sec:gemv_design_choices}

Our CUDA-core kernel addresses C1--C3 through the three design choices in Fig.~\ref{fig:cuda-core-kernel}(b).

\textbf{Multi-token CTA fusion (C1).}
We group multiple active decode tokens within a single cooperative thread array (CTA), i.e., a CUDA thread block, so each HBM-streamed weight tile is loaded once and reused across token-wise dot products without inflating $M$ to the 128-row Tensor-Core tile.

\textbf{Multi-accumulator threading (C2).}
We assign multiple output rows per thread, creating independent accumulator chains that the warp scheduler interleaves to expose instruction-level parallelism even when $M$ is small.

\textbf{Register-only dequantization (C3).}
We hold unpacked FP16 values exclusively in registers and feed them directly into \texttt{half2} FMAs, eliminating shared-memory materialization and the synchronization barriers it would require.
The resulting pipeline (Fig.~\ref{fig:cuda-core-kernel}(b)) overlaps load, conversion, scale application, and accumulation across successive $K$ tiles.

Together, these choices avoid the small-$M$ inefficiency of Tensor-Core GEMM by matching the actual decode shape.
Although the CUDA-core path dequantizes NVFP4 operands and performs half-precision FMA rather than native FP4 arithmetic, this decode-specialized implementation achieves up to $2.5\times$ lower projection latency than the NVFP4 Tensor-Core baseline (Table~\ref{tab:gemv_kernel}).
Implementation details and ablations are provided in Appendix~\ref{sec:appendix-kernel-details}.

%% file: Sections/experiments.tex
\vspace{-1mm}
\section{Experiments}
\label{sec:exp_settings}
\vspace{-1mm}
We evaluate the two axes of Sec.~\ref{sec:nvfp4_lrm}--\ref{sec:2.3} in turn: Sec.~\ref{sec:experimental_settings}--\ref{sec:results_reasoning} measure reasoning accuracy against PTQ baselines, Sec.~\ref{sec:kernels:results} measures decode-phase latency from kernel up to end-to-end, and Sec.~\ref{sec:ablation} ablates the policy components.

\subsection{Experimental Settings}
\label{sec:experimental_settings}

We evaluate ReSET on five LRMs from two families: R1-Distill-Qwen-7B/14B~\cite{deepseek-r1} and Qwen3-8B/14B/32B~\cite{qwen3}.
All models are quantized from public BF16 checkpoints to \emph{real} NVFP4 W4A4 with NVIDIA ModelOpt (E2M1 elements, E4M3 block scales, group size 16); the KV cache remains in BF16.
All baselines share this weight format and KV path on identical hardware, ensuring a like-for-like comparison.
We report averages over 8 seeds on AIME-120 (AIME 2022--2025 combined)~\cite{aime}, GPQA-Diamond~\cite{rein2023gpqagraduatelevelgoogleproofqa}, and LiveCodeBench~\cite{jain2024livecodebench}, using top-$p=0.95$ and \texttt{max\_tokens}=32k.
Baselines (RTN, BRQ~\cite{blockrotation}, 4/6~\cite{fouroversix}, MR-GPTQ~\cite{mrgptq}) decode at the model-default $T$=$0.6$; calibration protocols are in Appendix~\ref{app:evaluation_details}.
ReSET is applied on top of RTN.

\vspace{-2mm}
\subsection{Results on Reasoning Benchmarks}
\label{sec:results_reasoning}

\input{Tables/ptq_baselines}
\input{Tables/table_234}
\textbf{ReSET dominates the NVFP4 PTQ frontier.}
ReSET attains the best average accuracy on every benchmark in Table~\ref{tab:ptq_baselines}, surpassing all PTQ baselines, with the largest gain on AIME-120 (+2.6 over the NVFP4 baseline).
The improvement traces to the step-aware threshold of Sec.~\ref{sec:method}: the global threshold $\tau_0$ continues to identify symbolic tokens in low-uncertainty steps, while the step-relative threshold $\hat{H}_{\mathrm{step}}(t)$ rescues symbolic tokens whose absolute entropy is elevated by surrounding context.

\textbf{Step-aware control beats token-level heuristics.}
Table~\ref{tab:aime_temperature_scaling} compares different entropy-aware temperature policies on AIME-120 under the same decoding setup, where the primary difference lies in how entropy thresholds are determined.
Fixed-threshold and sliding-window baselines improve over NVFP4 inconsistently across models, whereas ReSET is best on every model, confirming that step-relative entropy is the operative signal.

\subsection{System Evaluation}
\label{sec:kernels:results}

\textbf{Kernel-level latency.}
Table~\ref{tab:gemv_kernel} compares our CUDA-core GEMV against vLLM's CUTLASS GEMM at decode batch $M{=}1$ on B200, the regime Sec.~\ref{sec:implementation} targets.
Across 12 projection shapes (QKV/O/Gate-Up/Down for Qwen3-8B/14B/32B), our kernel is \textbf{1.57--2.49$\times$} faster, with the largest gain on the Qwen3-8B down-projection ($N{=}4096$, $K{=}12288$).
Larger-$M$ and prefill Tensor-Core results, with per-choice ablations, are in Appendix~\ref{sec:appendix-kernel-details}.

\textbf{Quantization overhead is reduced, not added.}
A common concern with PTQ-time interventions is that they impose a computational burden on the decoding step.
Rotation-fused activation quantization (Table~\ref{tab:mrgptq_rotation_overhead}) costs $4.55$--$5.08\,\mu$s per layer at $M{=}1$ on Qwen3-32B, averaging about $2.3\!\times$ the standard activation-quantization cost (1.81--2.80\,$\mu$s).
Our pipeline further reduces this overhead by consuming a non-swizzled scale layout for the CUDA-core kernel, keeping activation quantization at $1.65$--$1.81\,\mu$s.
ReSET is a sampling-side intervention with no rotation, so the system inherits this lower pre-GEMM cost by construction.

\begin{figure}[t]
\centering
\centerline{\includegraphics[width=\columnwidth]{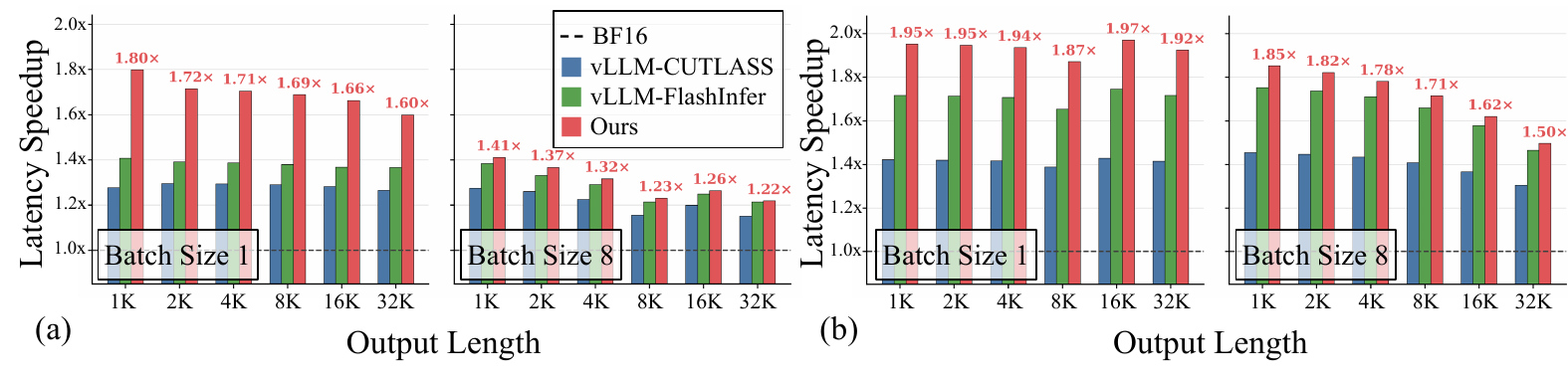}}
\vspace{-3mm}
\caption{
E2E speedup over BF16 for (a) Qwen3-8B and (b) Qwen3-32B with 512-token inputs.
}
\label{fig:e2e_speedup}
\vspace{-5mm}
\end{figure}

\textbf{End-to-end (E2E) latency reduction.}
\label{exp:e2e}
Fig.~\ref{fig:e2e_speedup} reports E2E speedup over BF16 on B200 (sm\_100a, vLLM + FlashInfer/CUTLASS, CUDA 13.1, PyTorch 2.10.0; details in Appendix~\ref{app:exp_details}) at \texttt{input\_len}=512, $B\in\{1,8\}$, and outputs up to 32k tokens.
Combining the NVFP4 kernels (Sec.~\ref{sec:implementation}) with the ReSET sampler (Sec.~\ref{sec:method}), \textbf{Ours} reaches \textbf{1.69$\times$} over BF16 and \textbf{1.22$\times$} over the strongest NVFP4 baseline on Qwen3-8B at $B{=}1$, output 8k; on Qwen3-32B, \textbf{1.97$\times$} ($B{=}1$) and \textbf{1.85$\times$} ($B{=}8$).
The ReSET sampler adds $\sim$100\,$\mu$s per generated token on Qwen3-32B---$\sim$1.5\% of a decode step---so the reported speedups already include the full sampling cost.

\subsection{Ablation Studies}
\label{sec:ablation}
We ablate the two policy choices below; sweeps for $w$, $T_{\mathrm{high}}$, $T_{\mathrm{low}}$, applying ReSET on top of GPTQ and Wikitext-2 perplexity across PTQ methods are in Appendix~\ref{app:ablation}.

\input{Tables/topp_sweep}
\textbf{Truncation-based sampling cannot substitute for step-aware control.}
With $T=0.6$ fixed, sweeping top-$p$ and min-$p$ thresholds (Table~\ref{tab:sampling_comparison}) yields limited or negative gains under NVFP4: top-$p$ improves the average only marginally, and min-$p$ consistently underperforms the NVFP4 baseline.
ReSET performs best across all models, confirming that step-aware entropy control — rather than distribution truncation — suppresses NVFP4's sampling errors.

\input{Tables/low_step_policy}
\textbf{Selective sharpening within low-uncertainty steps.}
Applying $T_{\mathrm{low}}$ uniformly to every token in low-uncertainty steps degrades accuracy (60.6 vs.\ 62.5 on AIME-120; Table~\ref{tab:low_step_policy}): not every such token is strictly deterministic, and uniform sharpening suppresses valid alternatives.
ReSET sharpens only the tokens below $\tau_0$ within those steps, leaving the rest unchanged.

\textbf{Hyperparameters are robust.}
Across reasonable ranges, ReSET's accuracy varies little: the window size $w\in\{16,32,64,128\}$ moves the AIME-120 average by 1.1 points (62.3 / 62.5 / 61.9 / 61.4; Table~\ref{tab:window_size}), and $T_{\mathrm{low}}\in\{0.1,0.2,0.3,0.4\}$ by 0.4 points (61.2--61.6; Table~\ref{tab:tlow_sweep}); $T_{\mathrm{high}}$ improves accuracy on every model as it grows from 0.6 toward 1.0, where the average peaks (Table~\ref{tab:thigh_sweep_aime120}).
Combined with the per-model entropy-percentile calibration of $\tau_0$ on five NuminaMath~\cite{numinamath} problems (Table~\ref{tab:tau_values}), ReSET deploys without task-specific tuning.

\vspace{-1mm}

%% file: Tables/ptq_baselines.tex
\begin{table}[t]
\centering
\caption{Comparison with NVFP4 PTQ methods across reasoning benchmarks.}
\label{tab:ptq_baselines}
\small

\resizebox{0.9\columnwidth}{!}{
\begin{tabular}{ll|ccccc|c}
\hline
Task & Method
& \makecell{R1-Qwen\\7B}
& \makecell{R1-Qwen\\14B}
& \makecell{Qwen3\\8B}
& \makecell{Qwen3\\14B}
& \makecell{Qwen3\\32B}
& Avg \\
\hline

\multirow{6}{*}{AIME-120}
& BF16 Baseline & 45.7 & 57.4 & 70.4 & 76.1 & 75.8 & 65.1 \\
& RTN     & 39.6 & 52.4 & 62.5 & 70.4 & 74.4 & 59.9 \\
& BRQ     & 41.4 & 49.8 & 53.8 & 66.9 & 73.0 & 57.0 \\
& 4/6     & 41.1 & 53.1 & 64.0 & 70.1 & 74.8 & 60.6 \\
& MR-GPTQ & 39.6 & 50.6 & 65.2 & 71.0 & 73.3 & 60.0 \\
& \cellcolor{mycolor}\textbf{ReSET} 
& \cellcolor{mycolor}43.8 
& \cellcolor{mycolor}54.0 
& \cellcolor{mycolor}64.9 
& \cellcolor{mycolor}72.1 
& \cellcolor{mycolor}77.5 
& \cellcolor{mycolor}\textbf{62.5} \\
\hline

\multirow{6}{*}{GPQA-Diamond}
& BF16 Baseline & 48.3 & 58.4 & 54.4 & 61.3 & 64.1 & 57.3 \\
& RTN     & 47.1 & 53.7 & 50.7 & 57.4 & 62.8 & 54.3 \\
& BRQ     & 45.5 & 56.7 & 53.5 & 58.2 & 60.8 & 55.0 \\
& 4/6     & 44.6 & 55.4 & 51.9 & 57.6 & 64.6 & 54.8 \\
& MR-GPTQ & 43.4 & 57.4 & 52.4 & 60.9 & 60.6 & 54.9 \\
& \cellcolor{mycolor}\textbf{ReSET} 
& \cellcolor{mycolor}46.0 
& \cellcolor{mycolor}57.6 
& \cellcolor{mycolor}53.2 
& \cellcolor{mycolor}58.5 
& \cellcolor{mycolor}62.9 
& \cellcolor{mycolor}\textbf{55.6} \\
\hline

\multirow{6}{*}{LiveCodeBench}
& BF16 Baseline & 28.2 & 38.5 & 43.2 & 46.5 & 49.9 & 41.3 \\
& RTN     & 29.5 & 37.1 & 36.4 & 46.7 & 46.5 & 39.2 \\
& BRQ     & 27.6 & 37.7 & 36.2 & 39.0 & 45.1 & 37.1 \\
& 4/6     & 27.0 & 36.4 & 35.8 & 45.3 & 45.7 & 38.0 \\
& MR-GPTQ & 27.2 & 34.9 & 42.9 & 47.2 & 45.1 & 39.5 \\
& \cellcolor{mycolor}\textbf{ReSET} 
& \cellcolor{mycolor}28.4 
& \cellcolor{mycolor}37.9 
& \cellcolor{mycolor}42.1 
& \cellcolor{mycolor}46.1 
& \cellcolor{mycolor}46.7 
& \cellcolor{mycolor}\textbf{40.2} \\
\hline
\end{tabular}
}
\vspace{-2mm}
\end{table}

%% file: Tables/table_234.tex
\begin{table*}[t]
\centering

\begin{minipage}[t]{0.51\textwidth}
\centering

\captionof{table}{Impact of temperature scaling methods.}
\label{tab:aime_temperature_scaling}
\resizebox{\columnwidth}{!}{
\begin{tabular}{l|cc|ccc|c}
\hline
\multirow{2}{*}{AIME-120 (↑)}
& \multicolumn{2}{c|}{R1-Qwen}
& \multicolumn{3}{c|}{Qwen3}
& \multirow{2}{*}{Avg} \\
\cline{2-6}
& 7B & 14B & 8B & 14B & 32B & \\
\hline
BF16         & 45.7 & 57.4 & 70.4 & 76.1 & 75.8 & 65.1 \\
\hline
NVFP4        & 39.6 & 52.4 & 62.5 & 70.4 & 74.4 & 59.9 \\
Fixed Threshold  & 41.8 & 53.2 & 63.3 & 70.3 & 75.8 & 60.9 \\
Sliding Window& 42.6 & 51.3 & 63.3 & 70.6 & 76.4 & 60.8 \\
\rowcolor{mycolor}
\textbf{ReSET} & \textbf{43.8} & \textbf{54.0} & \textbf{64.9} & \textbf{72.1} & \textbf{77.5} & \textbf{62.5} \\
\hline
\end{tabular}
}

\vspace{0.1em}

\captionof{table}{Per-projection pipeline latency at $M{=}1$ on Qwen3-32B ($\mu$s, ↓).
(Q: Quantize; R: Rotate.)}
\label{tab:mrgptq_rotation_overhead}
\resizebox{\columnwidth}{!}{
\resizebox{0.94\linewidth}{!}{%
\begin{tabular}{l|cc|cc|cc}
\hline
\multirow[c]{2}{*}{\makecell[c]{Layer\\Type}}
& \multicolumn{2}{c|}{vLLM}
& \multicolumn{2}{c|}{MR-GPTQ}
& \multicolumn{2}{c}{\cellcolor{mycolor}\textbf{Ours}} \\
\cline{2-7}
& Q & GEMM & Q+R & GEMM & \cellcolor{mycolor}Q & \cellcolor{mycolor}GEMV \\
\hline
QKV     & 1.81 & 13.41 & 4.65 & 11.84 & \cellcolor{mycolor}\textbf{1.65} & \cellcolor{mycolor}\textbf{6.62}  \\
Out     & 2.04 & 12.40 & 4.87 & 11.90 & \cellcolor{mycolor}\textbf{1.81} & \cellcolor{mycolor}\textbf{7.73}  \\
Gate-Up & 1.83 & 25.89 & 4.55 & 25.90 & \cellcolor{mycolor}\textbf{1.67} & \cellcolor{mycolor}\textbf{13.25} \\
Down    & 2.80 & 33.26 & 5.08 & 33.87 & \cellcolor{mycolor}\textbf{1.79} & \cellcolor{mycolor}\textbf{15.00} \\
\hline
\end{tabular}%
}
}

\end{minipage}
\hfill
\hfill
\begin{minipage}[t]{0.47\textwidth}
\centering

\captionof{table}{Kernel latency at $M{=}1$ across Qwen3 projections ($\mu$s, ↓).
Rows follow QKV, Out, Gate/Up, and Down.}
\label{tab:gemv_kernel}
\resizebox{\columnwidth}{!}{
\begin{tabular}{l|rr|rr|r}
\hline
Qwen3 & $N$ & $K$
& vLLM & \cellcolor{mycolor}\textbf{Ours} & Speedup \\
\hline
\multirow{4}{*}{8B}
&  6144 &  4096 & 8.69 & \cellcolor{mycolor}4.28 & 2.03$\times$ \\
&  4096 &  4096 & 7.69 & \cellcolor{mycolor}3.75 & 2.05$\times$ \\
& 12288 &  4096 & 11.45 & \cellcolor{mycolor}6.40 & 1.79$\times$ \\
&  4096 & 12288 & 16.61 & \cellcolor{mycolor}6.66 & 2.49$\times$ \\
\hline
\multirow{4}{*}{14B}
&  7168 &  5120 & 10.79 & \cellcolor{mycolor}5.53 & 1.95$\times$ \\
&  5120 &  5120 & 9.08 & \cellcolor{mycolor}4.64 & 1.96$\times$ \\
& 17408 &  5120 & 14.93 & \cellcolor{mycolor}9.49 & 1.57$\times$ \\
&  5120 & 17408 & 23.77 & \cellcolor{mycolor}10.35 & 2.30$\times$ \\
\hline
\multirow{4}{*}{32B}
& 10240 &  5120 & 13.41 & \cellcolor{mycolor}6.62 & 2.03$\times$ \\
&  5120 &  8192 & 12.40 & \cellcolor{mycolor}7.73 & 1.60$\times$ \\
& 25600 &  5120 & 25.89 & \cellcolor{mycolor}13.25 & 1.96$\times$ \\
&  5120 & 25600 & 33.26 & \cellcolor{mycolor}15.00 & 2.22$\times$ \\
\hline
\end{tabular}
}
\end{minipage}
\vspace{-4mm}
\end{table*}

%% file: Tables/topp_sweep.tex
\begin{table}[t]
\centering
\small

\centering
\caption{Comparison with truncation-based sampling methods on AIME-120.}
\label{tab:sampling_comparison}
\resizebox{0.8\columnwidth}{!}{
\begin{tabular}{l|ccc|cc|ccc|c}
\hline
\multirow{2}{*}{Method} 
& \multirow{2}{*}{$T$} 
& \multirow{2}{*}{Top-$p$}
& \multirow{2}{*}{Min-$p$}
& \multicolumn{2}{c|}{R1-Qwen}
& \multicolumn{3}{c|}{Qwen3}
& \multirow{2}{*}{Avg} \\
\cline{5-6} \cline{7-9}
& & & & 7B & 14B & 8B & 14B & 32B & \\
\hline
NVFP4 Baseline
& 0.6 & 0.95 & 0.0
& 39.6 & 52.4 & 62.5 & 70.4 & 74.4 & 59.9 \\
\hline
\multirow{5}{*}{\makecell{NVFP4 +\\Top-$p$ Sweep}}
& \multirow{5}{*}{0.6}
& 0.80
& \multirow{5}{*}{0.0}
& 41.0 & 49.5 & 61.0 & 70.1 & 74.3 & 59.2 \\
& & 0.85 & 
& 39.7 & 50.6 & 63.3 & 70.2 & 74.5 & 59.7 \\
& & 0.90 & 
& 40.2 & 52.4 & 63.0 & 71.2 & 74.5 & 60.3 \\
& & 0.99 & 
& 38.4 & 53.9 & 63.6 & 69.6 & 75.2 & 60.1 \\
& & 1.00 & 
& 40.1 & 52.6 & 63.9 & 70.7 & 73.6 & 60.2 \\
\hline
\multirow{4}{*}{\makecell{NVFP4 +\\Min-$p$ Sweep}}
& \multirow{4}{*}{0.6}
& \multirow{4}{*}{0.95}
& 0.01
& 39.8 & 52.4 & 61.2 & 68.7 & 74.8 & 59.4 \\
& & & 0.03
& 40.3 & 49.7 & 60.6 & 70.0 & 74.5 & 59.0 \\
& & & 0.05
& 38.7 & 50.2 & 58.5 & 70.2 & 73.3 & 58.2 \\
& & & 0.10
& 41.0 & 51.2 & 57.6 & 69.8 & 73.3 & 58.6 \\
\hline
\rowcolor{mycolor}
ReSET
& ReSET & 0.95 & 0.0
& \textbf{43.8} & \textbf{54.0} & \textbf{64.9} & \textbf{72.1} & \textbf{77.5} & \textbf{62.5} \\
\hline
\end{tabular}
}
\vspace{-5mm}
\end{table}

%% file: Tables/low_step_policy.tex
\begin{wraptable}{r}{0.50\linewidth}
\vspace{-3mm}
\centering
\caption{\small{Low-entropy step policy effect (AIME-120).}}
\label{tab:low_step_policy}
\small

\resizebox{0.50\textwidth}{!}{
\begin{tabular}{l|cc|ccc|c}
\hline
\multirow{2}{*}{Policy}
& \multicolumn{2}{c|}{R1-Qwen}
& \multicolumn{3}{c|}{Qwen3}
& \multirow{2}{*}{Avg} \\
\cline{2-6}
& 7B & 14B & 8B & 14B & 32B & \\
\hline
All $T_{\mathrm{low}}$
& 40.6 & 51.2 & 65.1 & 72.0 & 74.1 & 60.6 \\
\rowcolor{mycolor}
ReSET
& 43.8 & 54.0 & 64.9 & 72.1 & 77.5 & \textbf{62.5} \\
\hline
\end{tabular}
}

\vspace{-1.0em}
\end{wraptable}

%% file: Sections/conclusion.tex
\section{Conclusion}

We presented ReSET, a step-aware temperature scaling method for improving NVFP4 LRMs. By analyzing uncertainty at the reasoning-step level, ReSET adaptively controls decoding temperature to better preserve symbolic reasoning accuracy under NVFP4. We further introduced a CUDA-core small-M NVFP4 kernel optimized for latency-critical decoding. Experimental results show that ReSET consistently improves reasoning accuracy while achieving substantial decoding speedups.

%% file: Sections/appendix.tex
\section{Microscaling Formats}
\label{sec:appendix-mx}

\begin{table}[h]
\centering
\resizebox{0.7\columnwidth}{!}{%
\begin{tabular}{lcccc}
\toprule
Format & Element Type & Block Scale & Global Scale & Block Size \\ \midrule
MXFP4  & FP4 (E2M1)   & FP8 (E8M0)  & --           & 32 \\
NVFP4  & FP4 (E2M1)   & FP8 (E4M3)  & FP32         & 16 \\ \midrule
\end{tabular}%
}
\caption{Configuration details of MXFP4 and NVFP4 formats.}
\label{tab:mx_configs}
\end{table}

Table~\ref{tab:mx_configs} lists the key configuration parameters of two representative microscaled FP4 formats, MXFP4 and NVFP4. 
Both formats encode individual elements using FP4 with an E2M1 layout, while relying on additional scaling factors to recover dynamic range. 
In this family of formats, values are reconstructed through a block-level scale shared across a group of elements, optionally combined with a global scaling factor.

\textbf{MXFP4.}
MXFP4 adopts a block size of 32 and uses an FP8 (E8M0) representation for the block-wise scale. 
This exponent-only scale enables a broad dynamic range while keeping metadata compact, and the larger block size further reduces scaling overhead. 

\textbf{NVFP4.}
NVFP4 employs a smaller block size of 16 and represents the block scale in FP8 (E4M3), supplemented by an additional FP32 global scale. 
The combination of finer-grained blocks and an explicit global factor improves stability of magnitude calibration, albeit with slightly increased overhead. 
This configuration is particularly suited for low-precision pipelines (e.g., W4A4), where both activations and weights are quantized and more precise scaling control is required during long-context decoding.

\section{Additional Experimental Results and Details}
\label{app:exp_details}

\subsection{Analysis Details}
\label{appendix:token_analysis_settings}
We analyzed the entropy characteristics of tokens produced during extended reasoning processes. Our data was derived from the R1-Qwen-14B model's outputs across 90 mathematical challenges from the 2022--2024 AIME datasets~\cite{aime}.
The resulting corpus consists of over 1.5 million individual tokens, each associated with its respective entropy metric. For the final visualization, we applied a frequency filter, selecting only those tokens that appeared 1,000 times or more. We then partitioned these frequent tokens into low-entropy and high-entropy categories to illustrate the contrast between deterministic and high-variance token generation.

\paragraph{Definition of a Reasoning Step.}
We define a \emph{reasoning step} as a contiguous segment of tokens corresponding to a coherent intermediate reasoning unit. 
Following~\cite{lightman2024letsverify,chen2025seal}, we approximate step boundaries using a delimiter rule, where segments are separated by double newline tokens (``\textbackslash n\textbackslash n'').

\begin{figure}[t]
\centering
\centerline{\includegraphics[width=\columnwidth]{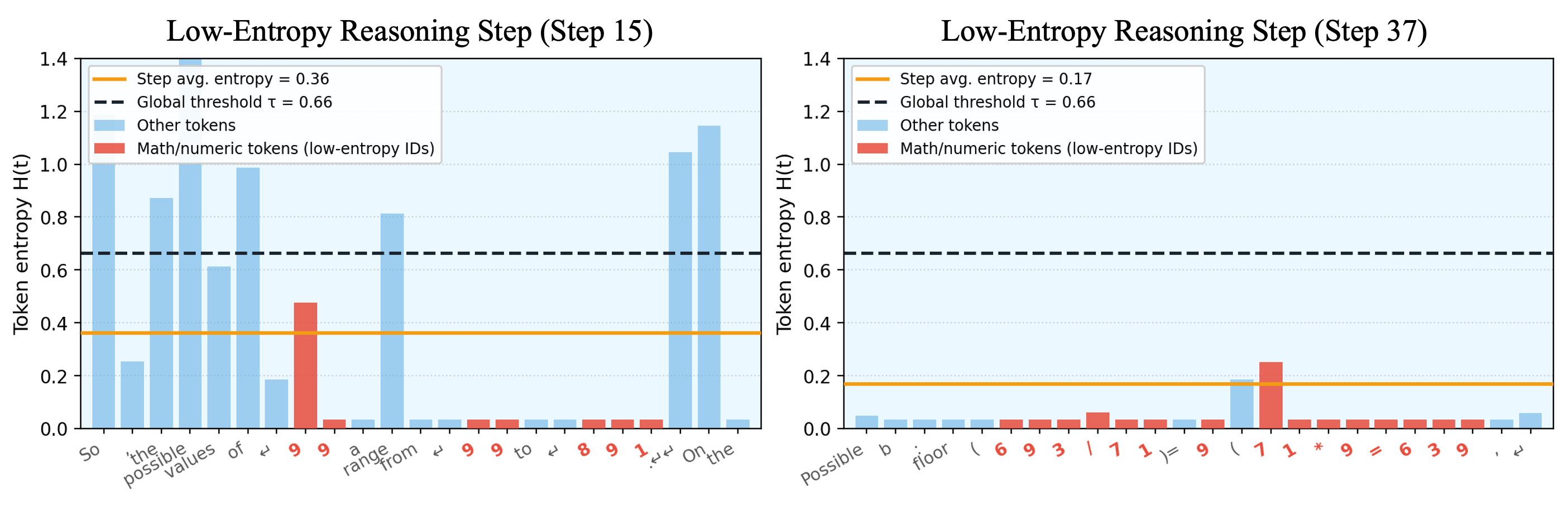}}
\caption{
Using $H_{step}$ only can misclassify low-entropy tokens.
}
\label{fig:fig_hres_failure}
\end{figure}

\begin{figure}[t]
\centering
\centerline{\includegraphics[width=\columnwidth]{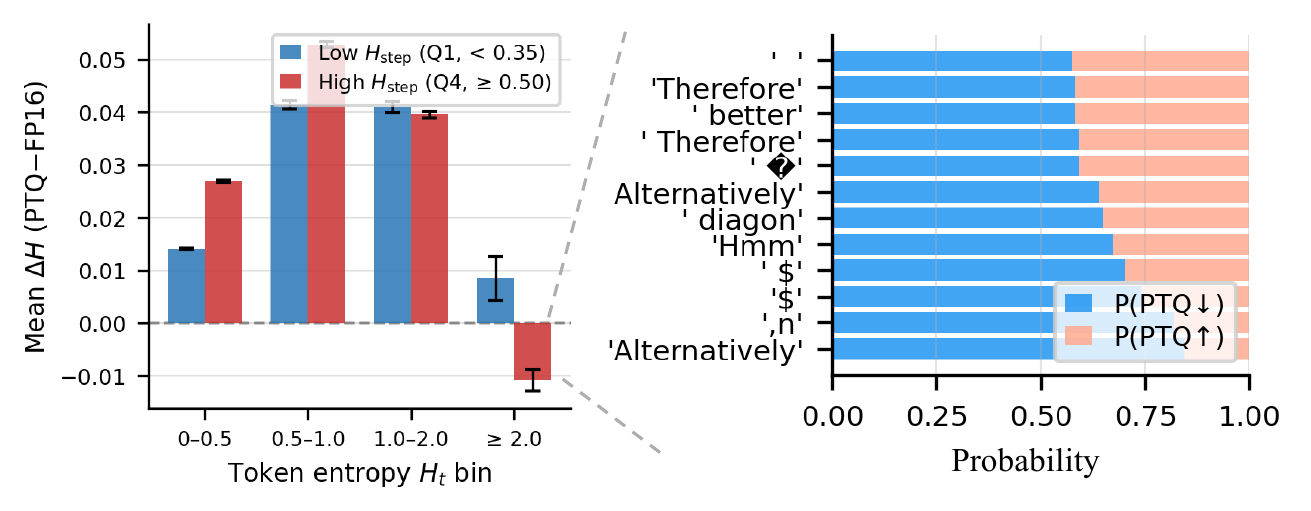}}
\caption{
Entropy shift under quantization.
(left) Mean entropy shift $\Delta H = H_{\mathrm{PTQ}}-H_{\mathrm{FP16}}$ across token entropy bins under low- and high-uncertainty steps.
(right) Probability of entropy increase or decrease for representative high-entropy tokens under PTQ.
}
\label{fig:threshold_comparison}
\end{figure}

\subsection{Quantization Distorts Entropy Differently Across Reasoning Steps}
\label{sec:quant_entropy_shift}

Sec.~\ref{sec:observations} shows that quantization increases sampling errors at low-entropy symbolic tokens by increasing the probability of non-top-1 alternatives.
However, we additionally observe that the direction of entropy change depends strongly on both token entropy and step-level uncertainty.

Fig.~\ref{fig:threshold_comparison} (left) shows the mean entropy shift $\Delta H_t = H_t^{\mathrm{PTQ}} - H_t^{\mathrm{FP16}}$ across token entropy bins under low- and high-uncertainty reasoning steps.
For low-entropy tokens, quantization consistently increases entropy, with larger shifts observed in high-uncertainty steps.
In contrast, for very high-entropy tokens ($H_t \ge 2.0$), the entropy shift becomes negative in high-uncertainty steps, indicating that quantization instead sharpens the token distribution.

At the token level (Fig.~\ref{fig:threshold_comparison} right), representative high-entropy discourse tokens such as ``Alternatively'' more frequently exhibit entropy reduction under PTQ.
This suggests that quantization can artificially concentrate probability mass on a smaller subset of continuations in uncertain reasoning states.

Overall, quantization introduces two distinct effects: entropy inflation for low-entropy symbolic tokens and entropy collapse for high-entropy tokens in uncertain reasoning steps.
These observations motivate step-aware decoding policies that adapt entropy thresholds according to the uncertainty of the current reasoning step.

\subsection{Step entropy estimators}
\begin{equation}
\begin{aligned}
\hat{H}_{\mathrm{step}}^{\mathrm{causal}}(t)
&=
\frac{1}{t - t_0 + 1}
\sum_{i=t_0}^{t} H_i, \\
\hat{H}_{\mathrm{step}}^{\mathrm{window}}(t)
&=
\frac{1}{|\mathcal{W}_t|}
\sum_{i \in \mathcal{W}_t} H_i, \\
\hat{H}_{\mathrm{step}}^{\mathrm{EMA}}(t)
&=
\alpha H_t + (1-\alpha)\hat{H}_{\mathrm{step}}^{\mathrm{EMA}}(t-1).
\end{aligned}
\label{eq:step_entropy_estimators}
\end{equation}

We provide the formal definitions of the step-entropy estimators used in Sec.~\ref{sec:method}. 
The causal estimator computes the running average over tokens within the current step, relying only on past observations. 
The sliding-window estimator restricts this average to a fixed-size window $\mathcal{W}_t$ to emphasize recent tokens. 
The exponential moving average (EMA) instead applies temporal smoothing with decay factor $\alpha$=0.1.

\subsection{Hardware and Software Settings}
\label{app:hardware_settings}

All experiments are performed on a single NVIDIA B200 GPU (sm\_100a, 148 SMs, 180\,GB HBM3e, 8\,TB/s peak HBM bandwidth) with CUDA 13.1 and PyTorch 2.10.0.
Inference is served through \texttt{vLLM} with the \texttt{FlashInfer-CUTLASS} backend.
Our NVFP4 runtime integrates two custom kernel variants: a CUDA-core GEMV variant for latency-critical small-$M$ decode and a Tensor-Core GEMM variant for larger-$M$ decode and prefill.
The GEMV and GEMM variants are registered as \texttt{torch.ops.nvfp4r.gemv} and \texttt{torch.ops.nvfp4r.gemm}, respectively.
During serving, the runtime dispatches to the GEMV variant for small-$M$ decode projections and to the GEMM variant when Tensor-Core tiles are sufficiently occupied.

\subsection{Evaluation Details}
\label{app:evaluation_details}
We conduct our experiments using the vLLM framework and models including Qwen3 and DeepSeek-R1, which are licensed under Apache 2.0 and MIT licenses, respectively. We ensure full compliance with these licenses by providing proper citations and adhering to their terms of use.

\textbf{NVFP4 Quantization.} 
We use the NVIDIA ModelOpt~\cite{modelopt} library for NVFP4 quantization.
All Transformer decoder layers are quantized.
Weights are calibrated using min--max quantization, and the global scale is calibrated on the C4 dataset~\cite{c4} using 32 samples with sequence length 2048.

\textbf{4/6 (Four-Over-Six).} 
We follow the original 4/6 algorithm, which selects quantization parameters by minimizing activation MSE, while weights are quantized using ModelOpt. 
For a fair comparison, we use a fixed global scale obtained from ModelOpt calibration instead of dynamic global scaling. 
The global scale is calibrated on the C4 dataset with sequence length 2048 and 32 samples.

\textbf{MR-GPTQ.} 
We follow the official MR-GPTQ implementation. 
Weight quantization is performed by minimizing MSE using the FineWeb-Edu dataset (sequence length 2048, 128 samples). 
Activation ranges are calibrated using MinMax. 
We disable the randomized Hadamard transform (block size 16), as it degrades accuracy under NVFP4 for Qwen3 models~\cite{mrgptq}.

\textbf{BRQ.} 
We implement BRQ with a randomized Hadamard transform using block size 16.

\textbf{Sampling setup.}
Unless otherwise stated, all results are obtained using stochastic decoding with temperature $0.6$ and top-$p$ sampling with $p=0.95$.

\input{Tables/tau_calib}
\textbf{Calibration for global threshold $\tau$}
We calibrate the global threshold $\tau$ using a small subset of the NuminaMath-1.5~\cite{numinamath} dataset.
Specifically, we randomly sample 5 problems, generate model responses, and compute token-level entropy for all generated tokens.
We then set $\tau$ as the 80th percentile of the resulting entropy distribution. $\tau$ used in our experiments are in Table~\ref{tab:tau_values}.

\subsection{Ablation Studies}
\label{app:ablation}
\input{Tables/window_sweep}
\textbf{Effect of sliding window size in step entropy estimation.}
We evaluate the impact of the window size $w$ used in the hybrid step-entropy estimator.
As shown in Table~\ref{tab:window_size}, performance is relatively stable across different window sizes, indicating that ReSET is not sensitive to the precise choice of $w$.
The best performance is achieved at $w=32$, which we use in all experiments.
Larger window sizes do not provide additional gains and can slightly degrade performance, suggesting that overly long windows blur step-local statistics and weaken the effectiveness of step-aware control.

\input{Tables/thigh_sweep}
\textbf{Effect of high-temperature scaling in uncertain steps.}
To validate the design choice of using a higher temperature in high-uncertainty steps, we evaluate ReSET with fixed $T_{\mathrm{low}}=0.1$ while sweeping $T_{\mathrm{high}}$. Table~\ref{tab:thigh_sweep_aime120} shows that increasing $T_{\mathrm{high}}$ consistently improves performance over the NVFP4 baseline across models. In particular, moving from the base temperature ($T_{\mathrm{high}}=0.6$) to higher values (0.8--1.0) yields substantial gains, indicating that higher temperature is beneficial in uncertain reasoning regimes. This trend is consistent with the observation in Sec.~\ref{sec:quant_entropy_shift} that quantization reduces entropy and concentrates token distributions in high-uncertainty steps. Increasing $T_{\mathrm{high}}$ counteracts this effect by restoring uncertainty, allowing the model to maintain multiple plausible continuations during reasoning.

\input{Tables/perplexity_comparison}
\textbf{Perplexity evaluation across PTQ methods.}
We evaluate Wikitext-2~\cite{merity2016pointerwiki} perplexity across PTQ methods in Table~\ref{tab:perplexity_comparison}. Advanced PTQ methods such as rotation and GPTQ provide only marginal perplexity gains over RTN under NVFP4, with differences remaining minimal across models.

\input{Tables/reset_gptq}
\textbf{Employing ReSET with GPTQ.}
As shown in Table~\ref{tab:reset_gptq}, while ReSET itself substantially improves over the vanilla RTN baseline across most tasks and model scales, additionally combining it with GPTQ provides little to no further benefit. In many settings, ReSET$^\dagger$ achieves performance comparable to, or only marginally different from, ReSET alone. We hypothesize that this behavior arises because conventional PTQ methods such as GPTQ mainly contribute by slightly reducing NVFP4 quantization errors, which indirectly helps control the increase of non-top1 token probabilities during low-entropy token selection, thereby improving reasoning accuracy. However, ReSET already performs a similar corrective role through its temperature scaling mechanism, effectively stabilizing the token distribution without requiring additional reconstruction-based PTQ refinement.

\input{Tables/tlow_sweep}

\section{NVFP4 Kernel Implementation Details}
\label{sec:appendix-kernel-details}

This appendix provides implementation details for the NVFP4 kernels used in Section~\ref{sec:implementation}.
We first describe the CUDA-core small-$M$ decode kernel, then describe the Tensor-Core GEMM path used for larger $M$ and prefill.
All kernel latencies are measured on a single NVIDIA B200 with CUDA-graph replay and are reported in microseconds.

\subsection{CUDA-Core Small-$M$ Decode Kernel}
\label{sec:appendix-gemv}

\paragraph{Multi-token CTA fusion.}
In small M scenario, naive CUDA-core mapping assigns different decode tokens to independent CTAs.
This mapping is simple, but it repeatedly streams the same weight tile for each active token because all decode tokens share the same weight matrix.
For small $M$, this repeated weight traffic is expensive: the amount of reuse available across tokens is limited, and launching independent CTAs prevents the reuse that does exist from being exploited within the CTA.

Our kernel instead fuses multiple active decode tokens within a CTA.
Each CTA is assigned a tile of output rows and a small group of active decode tokens.
It then traverses the $K$ dimension once, loads a tile of the shared weight matrix, and reuses the loaded weight values across token-wise dot products for the active tokens in the CTA.
This reduces redundant weight loads compared with processing each token independently, while avoiding padding the token dimension to the fixed Tensor-Core tile size.
The approach is especially useful for small but non-unit decode batches, where $M$ is still too small for efficient Tensor-Core GEMM but large enough to benefit from intra-CTA weight reuse.

\paragraph{Multi-accumulator threading.}
After moving from Tensor Cores to CUDA cores, the projection is implemented as many \texttt{half2} multiply-accumulate operations.
If each thread maintains only one output accumulator, the inner loop forms a single dependent FMA chain along the $K$ dimension:
each new accumulation depends on the result of the previous \texttt{hfma2}.
This limits instruction-level parallelism and can leave CUDA cores underutilized, especially when the loop also includes FP4 unpacking and scale application.

To reduce this dependency bottleneck, each thread maintains multiple output accumulators.
In our evaluated configuration, each thread owns two weight rows and accumulates two independent output values.
The two accumulators create independent \texttt{hfma2} streams, allowing the warp scheduler to interleave instructions from different dependency chains.
This hides part of the latency from conversion, scaling, and FMA instructions without increasing shared-memory traffic.
The number of accumulators is chosen to balance instruction-level parallelism against register pressure: using too few accumulators exposes insufficient parallelism, while using too many reduces occupancy by increasing per-thread register usage.

\paragraph{Register-only dequantization.}
CUDA-core NVFP4 execution must explicitly perform the unpacking and scaling operations that are normally handled inside Tensor-Core MMA.
Packed FP4 values are loaded into registers, converted to FP16 pairs using \texttt{cvt.rn.f16x2.e2m1x2}, multiplied by the corresponding shared NVFP4 scale, and then consumed by \texttt{hfma2} accumulation.
The activation values and their scales are handled similarly, so the inner loop performs dequantization and accumulation together.

A straightforward implementation could first materialize the dequantized FP16 values in shared memory and then load them again for accumulation.
However, this introduces additional shared-memory traffic, increases synchronization requirements, and lengthens the inner loop.
These costs are particularly harmful in the small-$M$ regime, where the compute work per CTA is limited and fixed overheads are difficult to amortize.

Our kernel instead keeps the converted FP16 values, scales, and partial accumulators in registers.
Conversion and scale application are scheduled close to the consuming \texttt{hfma2} instructions, so the dequantized values are produced and consumed without being written to shared memory.
Across successive $K$ tiles, the kernel overlaps weight loading, FP4 conversion, scale multiplication, and accumulation through software pipelining.
This register-only design reduces memory traffic and synchronization overhead while preserving the low memory footprint of NVFP4 weights.

\paragraph{Runtime dispatch.}
The CUDA-core kernel is used only in the small-$M$ decode regime, where Tensor-Core GEMM suffers from tile under-occupancy.
For larger $M$ and prefill, we dispatch to the Tensor-Core NVFP4 GEMM path, where the fixed Tensor-Core tile is sufficiently occupied and the higher peak throughput becomes beneficial.
The dispatch decision is made per shape and decode-batch regime, so the serving runtime can use the CUDA-core path for latency-critical decoding without affecting the high-throughput prefill path.

\subsection{CUDA-Core Kernel Ablation}
\label{sec:appendix-gemv-ablation}

\begin{table}[h]
\centering
\caption{CUDA-core small-$M$ kernel ablation on Qwen3-32B \texttt{down\_proj} ($N{=}5120$, $K{=}25600$). Latency is reported in microseconds, measured on B200 with CUDA-graph replay.}
\label{tab:gemv-ablation}
\begin{tabular}{r|ccc}
\toprule
                                  & $M=1$  & $M=4$  & $M=8$  \\ \midrule
Ours: full kernel (C1+C2+C3)      & 15.0 & 27.5 & 58.3 \\ \midrule
without C1 (no CTA fusion)        & 15.3 & 35.9 & 70.9 \\
without C2 (single-row threading) & 16.8 & 32.4 & 64.6 \\
without C1 and C2                 & 18.7 & 39.4 & 76.5 \\ \bottomrule
\end{tabular}
\end{table}

Table~\ref{tab:gemv-ablation} shows that CTA fusion and multi-accumulator threading address complementary bottlenecks.
CTA fusion becomes more important as $M$ grows, reducing latency from $70.9$\textmu s to $58.3$\textmu s at $M=8$ by reusing each streamed weight tile across multiple decode tokens.
Multi-accumulator threading improves all regimes, including $M\!=\!1$, by increasing per-thread instruction-level parallelism.
Removing both choices compounds the slowdown, increasing latency by $1.31\times$ at $M\!=\!8$.

We do not isolate register-only dequantization as a separate ablation because replacing it with shared-memory materialization changes the kernel pipeline itself, including shared-memory allocation, synchronization placement, and pipeline depth.
Instead, the ablation focuses on mapping choices that can be removed while preserving the same CUDA-core NVFP4 execution structure.
Register-only dequantization remains part of the full kernel because it avoids extra shared-memory traffic and enables overlap between conversion, scale application, and accumulation in the inner $K$ loop.

\subsection{Tensor-Core GEMM Kernel and Runtime Dispatch}
\label{sec:appendix-gemm}

\input{Tables/kernel-level_latency_M=48}
\input{Tables/kernel-level_latency_M=128}

In addition to the CUDA-core small-$M$ kernel, we implement an NVFP4 Tensor-Core GEMM kernel for larger $M$ and prefill.
This Tensor-Core kernel is separate from vLLM's \texttt{cutlass\_scaled\_fp4\_mm}, which we use as the baseline in this comparison.
It targets the NVFP4 layout and the Qwen3 linear-layer shapes evaluated in this work, allowing us to tune tiling, scheduling, and scale handling directly for the target workloads.

Tables~\ref{tab:kernel_M4_M8} and~\ref{tab:gemm_kernel} compare our kernels with vLLM's \texttt{cutlass\_scaled\_fp4\_mm} (vLLM-CUTLASS).
Table~\ref{tab:kernel_M4_M8} reports latency at small decode sizes ($M{=}4,8$).
Here, Ours denotes the faster of our CUDA-core small-$M$ kernel and our Tensor-Core GEMM kernel for each layer shape, and speedup is measured over vLLM-CUTLASS.
Table~\ref{tab:gemm_kernel} reports latency at $M{=}128$, where our Tensor-Core GEMM kernel is used; our kernel is up to $1.34\times$ faster than vLLM-CUTLASS.

The runtime integrates the CUDA-core small-$M$ kernel and the Tensor-Core GEMM kernel through a shape-aware dispatch policy.
For $M{=}1,2$, the CUDA-core small-$M$ kernel is always used because it is faster in this regime.
For $M\geq128$, the Tensor-Core GEMM kernel is used because the fixed Tensor-Core tile is sufficiently occupied.
For intermediate sizes, the faster kernel depends on the projection shape, so the runtime selects between the CUDA-core and Tensor-Core kernels based on measured latency for each $(M,\mathrm{shape})$ pair.
Because vLLM captures a separate CUDA graph for each batch size, this selection is fixed at graph-capture time and incurs no branching overhead during replay.

\section{Limitations}
\label{limitation}

Our analysis and evaluation focus primarily on reasoning-oriented autoregressive decoding workloads under NVFP4 quantization. Although ReSET consistently improves reasoning accuracy across the evaluated model families and benchmarks, the relationship between token entropy and quantization-induced reasoning degradation may vary across other architectures, domains, or task types that were not explored in this work. In particular, these entropy dynamics may depend on how the underlying reasoning model is trained, including factors such as post-training reinforcement learning, supervised fine-tuning objectives, or reasoning-trace distillation strategies. As a result, the optimal entropy thresholds and temperature hyperparameters may vary depending on model-specific entropy characteristics and decoding behavior.

In addition, our analysis attributes many reasoning failures to incorrect sampling at locally deterministic low-entropy tokens. While we observe that such symbolic-token errors frequently correlate with downstream reasoning failures, reasoning degradation in quantized models is likely influenced by multiple interacting factors beyond isolated symbolic mistakes alone. For example, quantization may also affect long-range reasoning consistency, discourse planning, or intermediate semantic representations in ways not fully captured by token-level entropy analysis.

    Our experiments also focus on NVFP4 inference on NVIDIA Blackwell GPUs and latency-critical small-batch decoding settings. Although this setting is practically important for modern reasoning-model serving, the effectiveness of the proposed decoding strategy and kernel design under other hardware platforms, larger-batch throughput-oriented serving regimes, or alternative low-precision formats remains underexplored. In particular, more aggressive microscaling formats such as MXFP4 may introduce larger quantization errors and different entropy dynamics, which could provide an important direction for future investigation.

\section{Broader Impacts}
This work studies low-precision inference for large reasoning models. Improving quantized inference efficiency may help reduce deployment cost and resource usage for large language models. However, low-precision quantization can also introduce unintended reasoning errors or incorrect generations, particularly in applications requiring reliable multi-step reasoning. Our work aims to mitigate accuracy degradations in reasoning benchmarks under NVFP4 inference, but quantized reasoning models may still exhibit unexpected behaviors depending on the deployment setting and quantization configuration.

\section{LLM Usage}
We employed large language models (LLMs) exclusively for the purpose of code refinement and grammatical editing of the manuscript. All scientific contributions, experimental results, and interpretations were generated and verified by the human authors.

%% file: Tables/tau_calib.tex
\begin{table}[h]
\centering
\caption{Global entropy threshold $\tau$ (80th percentile) per model.}
\label{tab:tau_values}
\small
\begin{tabular}{lc}
\toprule
Model & $\tau$ \\
\midrule
R1-7B & 0.6446 \\
R1-14B & 0.6488 \\
Qwen3-8B & 0.5505 \\
Qwen3-14B & 0.4863 \\
Qwen3-32B & 0.5363 \\
\bottomrule
\end{tabular}
\end{table}

%% file: Tables/window_sweep.tex
\begin{table}[t]
\centering
\caption{Sensitivity to sliding window size $w$ in step entropy estimation (AIME-120).}
\label{tab:window_size}
\small
\begin{tabular}{c|ccccc|c}
\toprule
$w$ 
& R1-Qwen-7B & R1-Qwen-14B & Qwen3-8B & Qwen3-14B & Qwen3-32B & Avg \\
\midrule
16  & 44.9 & 53.9 & 65.7 & 70.3 & 76.7 & 62.3 \\
32  & 43.8 & 54.0 & 64.9 & 72.1 & 77.5 & \textbf{62.5} \\
64  & 44.5 & 54.4 & 63.5 & 72.0 & 75.1 & 61.9 \\
128 & 43.2 & 52.0 & 64.7 & 71.8 & 75.3 & 61.4 \\
\bottomrule
\end{tabular}
\end{table}

%% file: Tables/thigh_sweep.tex
\begin{table}[t]
\centering
\caption{AIME-120 score under NVFP4 ReSET with $T_{\mathrm{low}}=0.1$ across different $T_{\mathrm{high}}$.}
\label{tab:thigh_sweep_aime120}
\small
\begin{tabular}{l|ccccc}
\toprule
$T_{\mathrm{high}}$ 
& R1-Qwen-7B & R1-Qwen-14B & Qwen3-8B & Qwen3-14B & Qwen3-32B \\
\midrule
0.6 (Baseline) & 39.6 & 50.9 & 61.9 & 71.4 & 74.9 \\
0.7            & 39.8 & 51.6 & 64.3 & 70.6 & 74.1 \\
0.8            & 40.4 & 54.8 & 64.5 & 73.7 & 74.9 \\
0.9            & 43.8 & 54.5 & 64.6 & 71.1 & 75.3 \\
1.0 (Default in ReSET) & 42.0 & 52.9 & 64.9 & 70.9 & 77.5 \\
\bottomrule
\end{tabular}
\end{table}

%% file: Tables/perplexity_comparison.tex
\begin{table}[t]
\centering
\caption{Perplexity Comparison across Models}
\label{tab:perplexity_comparison}
\small
\begin{tabular}{ll|ccccc}
\toprule
Bit-Precision & Method 
& R1-Qwen-7B & R1-Qwen-14B & Qwen3-8B & Qwen3-14B & Qwen3-32B \\
\midrule

BF16 & Baseline 
& 25.07 & 8.91 & 9.72 & 8.65 & 7.61 \\
\cmidrule(lr){1-7}

\multirow{4}{*}{NVFP4}
& RTN     
& 26.23 & 9.50 & 10.10 & 8.91 & 8.01 \\
& 4/6     
& 26.19 & 9.49 & 10.08 & 8.86 & 7.98 \\
& BRQ     
& 27.00 & 9.81 & 10.35 & 9.53 & 8.17 \\
& MR-GPTQ 
& 27.48 & 9.64 & 10.06 & 8.95 & 7.81 \\

\bottomrule
\end{tabular}
\end{table}

%% file: Tables/reset_gptq.tex
\begin{table}[t]
\centering
\caption{Comparison with NVFP4 PTQ methods across tasks. $\dagger$: ReSET applied on top of GPTQ.}
\label{tab:reset_gptq}
\small

\resizebox{0.9\columnwidth}{!}{
\begin{tabular}{ll|ccccc|c}
\hline
Task & Method
& \makecell{R1-Qwen\\7B}
& \makecell{R1-Qwen\\14B}
& \makecell{Qwen3\\8B}
& \makecell{Qwen3\\14B}
& \makecell{Qwen3\\32B}
& Avg \\
\hline

\multirow{3}{*}{AIME-120}
& RTN
& 39.6 & 52.4 & 62.5 & 70.4 & 74.4 & 59.9 \\
& \cellcolor{mycolor}\textbf{ReSET}
& \cellcolor{mycolor}43.8
& \cellcolor{mycolor}54.0
& \cellcolor{mycolor}64.9
& \cellcolor{mycolor}72.1
& \cellcolor{mycolor}77.5
& \cellcolor{mycolor}\textbf{62.5} \\
& \cellcolor{mycolor}\textbf{ReSET$^\dagger$}
& \cellcolor{mycolor}41.9
& \cellcolor{mycolor}53.5
& \cellcolor{mycolor}65.0
& \cellcolor{mycolor}72.4
& \cellcolor{mycolor}76.1
& \cellcolor{mycolor}\uline{61.6} \\
\hline

\multirow{3}{*}{GPQA-Diamond}
& RTN
& 47.1 & 53.7 & 50.7 & 57.4 & 62.8 & 54.3 \\
& \cellcolor{mycolor}\textbf{ReSET}
& \cellcolor{mycolor}46.0
& \cellcolor{mycolor}57.6
& \cellcolor{mycolor}53.2
& \cellcolor{mycolor}58.5
& \cellcolor{mycolor}62.9
& \cellcolor{mycolor}\textbf{55.6} \\
& \cellcolor{mycolor}\textbf{ReSET$^\dagger$}
& \cellcolor{mycolor}44.4
& \cellcolor{mycolor}57.4
& \cellcolor{mycolor}52.2
& \cellcolor{mycolor}59.6
& \cellcolor{mycolor}64.3
& \cellcolor{mycolor}\textbf{55.6} \\
\hline

\multirow{3}{*}{LiveCodeBench}
& RTN
& 29.5 & 37.1 & 36.4 & 46.7 & 46.5 & 39.2 \\
& \cellcolor{mycolor}\textbf{ReSET}
& \cellcolor{mycolor}28.4
& \cellcolor{mycolor}37.9
& \cellcolor{mycolor}42.1
& \cellcolor{mycolor}46.1
& \cellcolor{mycolor}46.7
& \cellcolor{mycolor}\uline{40.2} \\
& \cellcolor{mycolor}\textbf{ReSET$^\dagger$}
& \cellcolor{mycolor}29.5
& \cellcolor{mycolor}37.3
& \cellcolor{mycolor}43.8
& \cellcolor{mycolor}46.7
& \cellcolor{mycolor}48.8
& \cellcolor{mycolor}\textbf{41.2} \\
\hline

\end{tabular}
}
\end{table}

%% file: Tables/tlow_sweep.tex
\begin{table}[t]
\centering
\caption{Sensitivity to $T_{\mathrm{low}}$ with fixed $T_{\mathrm{high}}=1.0$ in ReSET (AIME-120).}
\label{tab:tlow_sweep}
\small
\begin{tabular}{c|ccccc|c}
\toprule
$T_{\mathrm{low}}$ 
& R1-Qwen-7B & R1-Qwen-14B & Qwen3-8B & Qwen3-14B & Qwen3-32B & Avg \\
\midrule
0.1 & 42.0 & 52.9 & 64.9 & 70.9 & 77.5 & \textbf{61.6} \\
0.2 & 42.7 & 52.9 & 64.3 & 70.7 & 75.7 & 61.2 \\
0.3 & 41.4 & 54.0 & 64.1 & 72.1 & 75.6 & 61.4 \\
0.4 & 43.8 & 52.2 & 64.8 & 71.2 & 75.4 & 61.5 \\
\bottomrule
\end{tabular}
\end{table}

%% file: Tables/kernel-level_latency_M=48.tex
\begin{table}[t]
\centering
\caption{Per-layer latency at small decode sizes ($M{=}4,8$). Latency is reported in microseconds.}
\label{tab:kernel_M4_M8}
\small
\setlength{\tabcolsep}{4pt}
\begin{tabular}{@{}cccccccccc@{}}
\toprule
& & & &
\multicolumn{3}{c}{$M{=}4$} &
\multicolumn{3}{c}{$M{=}8$} \\
\cmidrule(lr){5-7}\cmidrule(lr){8-10}
\textbf{Model} & \textbf{Layer} & $N$ & $K$
& \textbf{vLLM} & \textbf{Ours} & \textbf{Speedup}
& \textbf{vLLM} & \textbf{Ours} & \textbf{Speedup} \\
\midrule
Qwen3-8B  & qkv\_proj     &  6144 &  4096 &  8.32 &  6.18 & $1.35\times$ &  8.05 &  6.01 & $1.34\times$ \\
Qwen3-8B  & o\_proj       &  4096 &  4096 &  7.47 &  6.00 & $1.25\times$ &  7.42 &  5.81 & $1.28\times$ \\
Qwen3-8B  & gate/up\_proj & 12288 &  4096 & 10.34 &  8.67 & $1.19\times$ & 10.35 &  8.26 & $1.25\times$ \\
Qwen3-8B  & down\_proj    &  4096 & 12288 & 16.05 & 11.57 & $1.39\times$ & 15.49 & 11.62 & $1.33\times$ \\
\addlinespace
Qwen3-14B & qkv\_proj     &  7168 &  5120 & 10.92 &  7.40 & $1.48\times$ & 10.09 &  7.42 & $1.36\times$ \\
Qwen3-14B & o\_proj       &  5120 &  5120 &  9.09 &  6.88 & $1.32\times$ &  8.66 &  6.81 & $1.27\times$ \\
Qwen3-14B & gate/up\_proj & 17408 &  5120 & 14.88 & 11.21 & $1.33\times$ & 13.59 & 11.19 & $1.21\times$ \\
Qwen3-14B & down\_proj    &  5120 & 17408 & 23.58 & 15.40 & $1.53\times$ & 22.22 & 15.02 & $1.48\times$ \\
\addlinespace
Qwen3-32B & qkv\_proj     & 10240 &  5120 & 13.50 &  8.43 & $1.60\times$ & 11.95 &  8.45 & $1.41\times$ \\
Qwen3-32B & o\_proj       &  5120 &  8192 & 12.31 &  9.31 & $1.32\times$ & 11.92 &  9.46 & $1.26\times$ \\
Qwen3-32B & gate/up\_proj & 25600 &  5120 & 26.08 & 17.51 & $1.49\times$ & 23.14 & 17.33 & $1.34\times$ \\
Qwen3-32B & down\_proj    &  5120 & 25600 & 34.14 & 20.54 & $1.66\times$ & 31.94 & 20.56 & $1.55\times$ \\
\bottomrule
\end{tabular}
\end{table}

%% file: Tables/kernel-level_latency_M=128.tex
\begin{table}[t]
\centering
\caption{Per-layer latency for NVFP4 Tensor-Core GEMM at $M{=}128$. Latency is reported in microseconds.}
\label{tab:gemm_kernel}
\small
\setlength{\tabcolsep}{5pt}
\begin{tabular}{@{}ccccccc@{}}
\toprule
\textbf{Model} & \textbf{Layer} & $N$ & $K$
& \textbf{vLLM} $(\mu s)$ & \textbf{Ours} $(\mu s)$ & \textbf{Speedup} \\
\midrule
Qwen3-8B  & qkv\_proj     &  6144 &  4096 & 13.76 & 11.36 & 1.21$\times$ \\
Qwen3-8B  & o\_proj       &  4096 &  4096 & 13.79 & 11.23 & 1.23$\times$ \\
Qwen3-8B  & gate/up\_proj & 12288 &  4096 & 18.85 & 14.78 & 1.27$\times$ \\
Qwen3-8B  & down\_proj    &  4096 & 12288 & 18.91 & 14.82 & 1.28$\times$ \\
\addlinespace
Qwen3-14B & qkv\_proj     &  7168 &  5120 & 13.89 & 11.30 & 1.23$\times$ \\
Qwen3-14B & o\_proj       &  5120 &  5120 & 13.79 & 11.52 & 1.20$\times$ \\
Qwen3-14B & gate/up\_proj & 17408 &  5120 & 20.93 & 18.69 & 1.12$\times$ \\
Qwen3-14B & down\_proj    &  5120 & 17408 & 25.06 & 18.88 & 1.33$\times$ \\
\addlinespace
Qwen3-32B & qkv\_proj     & 10240 &  5120 & 18.91 & 14.94 & 1.27$\times$ \\
Qwen3-32B & o\_proj       &  5120 &  8192 & 16.86 & 12.74 & 1.32$\times$ \\
Qwen3-32B & gate/up\_proj & 25600 &  5120 & 25.06 & 24.99 & 1.00$\times$ \\
Qwen3-32B & down\_proj    &  5120 & 25600 & 33.25 & 24.80 & 1.34$\times$ \\
\bottomrule
\end{tabular}
\end{table}